\documentclass[sigconf]{acmart}

\usepackage{booktabs} 

\usepackage[english]{babel}
\usepackage{moresize}
\usepackage{amsmath}

\usepackage{algorithmic} 
\usepackage{balance}
\usepackage{comment}
\usepackage{paralist}

\usepackage[english]{babel}
\usepackage[latin1]{inputenc}
\usepackage{mathrsfs}
\usepackage{graphicx}
\usepackage{amssymb}
\usepackage{url}
\usepackage{mathrsfs}
\usepackage{subfigure}
\usepackage{amsmath}
\usepackage{enumitem}
\usepackage[linesnumbered,algoruled,boxed,lined]{algorithm2e}
\usepackage{adjustbox}
\usepackage{amssymb}
\usepackage{times}

\newcommand{\hide}[1]{} 

\newcommand{\etal}{\textit{et al}.}

\newcommand{\red}[1]{{\color{red}#1}}
\newcommand{\ie}{\textit{i}.\textit{e}.}
\newcommand{\eg}{\textit{e}.\textit{g}.} 
\newcommand{\wrt}{\textit{w}.\textit{r}.\textit{t}}

\DeclareMathOperator*{\argmin}{argmin}
\setcopyright{none}

\acmConference[KDD'18]{International Conference of Knowledge Discovery and Data Mining}{Aug 2018}{London, UK} 
\acmYear{2018}
\copyrightyear{2017}

\def\model{NTF}
\def\full{Neural network based Tensor Factorization}

\begin{document}
\title{Neural Tensor Factorization}



\author{Xian Wu$^1$\quad Baoxu Shi$^1$\quad Yuxiao Dong$^2$\quad Chao Huang$^1$\quad Nitesh V. Chawla$^1$}
\affiliation{University of Notre Dame$^1$, \quad Microsoft Research$^2$\\
    \{xwu9, bshi, chuang7, nchawla\}@nd.edu\quad yuxdong@microsoft.com}

%
%
%
%
%
%

\renewcommand{\authors}{Xian Wu, Baoxu Shi, Yuxiao Dong, Chao Huang, Nitesh V. Chawla}
\renewcommand{\shortauthors}{X. Wu et al.}

\begin{abstract}


 Neural collaborative filtering (NCF)~\cite{he2017neural} and recurrent recommender systems (RRN)~\cite{wu2017recurrent} have been successful in modeling user-item relational data. However, they are also limited in their assumption of static or sequential modeling of relational data as they  do not account for evolving users' preference over time as well as changes in the underlying factors that drive the change in user-item relationship over time. We address these limitations by proposing a  \full\ (\model) model for predictive tasks on dynamic relational data. 
The \model\ model generalizes conventional tensor factorization from two perspectives: 
First, it leverages the long short-term memory architecture to  characterize the multi-dimensional temporal interactions on relational data. 
Second, it  incorporates the multi-layer perceptron structure for learning the non-linearities between different latent factors. 
Our extensive experiments demonstrate the significant improvement in rating prediction and link prediction on dynamic relational data by our \model\ model over both neural network based factorization models and other traditional methods.  

\hide{
While significant progress has been made to apply deep neural networks to various tasks (\eg, recommendations and predictive analysis) in relational data (\eg, user rating data), some important challenges have not been well addressed yet: (i) we recognize the importance of time dimension in modeling the interactions (\eg, user's rating) among relational data. (ii) The inherent factors which affect interactions may change over time. As a consequence, it is infeasible to model them using a fixed temporal pattern. In this paper, we represent dynamic relational data as tensors and model the time-evolving interactions between multi-dimensions by proposing a general framework \underline{\textbf{N}}eural network based \underline{\textbf{T}}ensor \underline{\textbf{F}}actorization (\model). \model\ generalizes tensor factorization framework with temporal dynamics by (i) modeling dynamic characteristics of multi-dimensional interactions with recurrent neural networks, and (ii) further integrating the temporal constrains into the process of tensor factorization in a non-linear manner. Finally, we perform extensive experiments to demonstrate the better ability of \model\ in a variety of applications, including rating prediction and link prediction, compared with state-of-the-art techniques.

}

\end{abstract}

%
%




\maketitle

\section{Introduction}
\label{sec:intro}

Learning from relational data (\eg, the user-item interactions on Netflix) has benefited many real-world services and applications, such as rating prediction and item recommendation on online platforms. 
A significant line of research has shown that latent factor models, in particular factorization based techniques, offer state-of-the-art results for learning tasks on relational data~\cite{Koren:2008KDD,mnih2008probabilistic,Ma:2008CIKM08}. 
In addition, as the main framework of the winning solution of the Netflix Prize, matrix  factorization (MF) has demonstrated its power on industrial-grade applications~\cite{Koren:2009MF}, further attracting much effort in generalizing its predictive abilities.

One direction of these efforts has been devoted to extending a two-dimensional matrix, representative of interactions between users and items, into a three-dimensional tensor for incorporating the time information.  
Subsequently, the tensor factorization (TF) technique can be employed to project users and items into a latent space with the encoding of time~\cite{kolda2009tensor,bhargava2015and}. 
However, conventional TF assumes the independence between two consecutive time slots, leaving it infeasible to make predictions for the next time slot. 
Further, it is also incapable of capturing the temporal patterns that are themselves time-evolving, 
such as  
(i) the fast-changing item perception, for example, individuals' impression to a movie may be dynamically affected by its winning of some movie awards. 
and  
(ii) the evolution of users' preferences, i.e., user's tastes may change over time.

Recently, another attempt---recurrent recommender networks (RRN)---was made to integrate a recurrent neural network  with factorization models for modeling the sequence dependencies between users' behavioral trajectories~\cite{wu2017recurrent}. 
However, RRN achieves this by setting a fixed length of items' historical ratings, ignoring the time interval between two consecutive ratings. 
Consequently, these fixed-length (\eg, $k$) rating sequences  may cover various time-frames for different items. This is a limitation of RRN. 
For example, a popular movie may only take one hour to receive $k$ ratings, while a cult movie may need days to collect the same count of ratings. 
Additionally, both RRN and conventional TF models utilize dot product to make rating predictions, missing the potential to model the nonlinearities between latent factors.

To address these challenges and limitations, we develop a  \full\ model (\model). 
In general, the \model\ takes a three-way tensor (i.e., user-item-time) as input and learns the latent embeddings (commonly referred to as factors in TF) for each dimension of the tensor. 
Specifically, \model\ integrates the long short-term memory (LSTM) network with tensor factorization. 
The LSTM module is used to adaptively capture the dependencies among multi-dimensional interactions based on the learned representations for each time slot. 
Furthermore, instead of using dot product between learned representations  to make rating predictions, \model\ concatenates the inherent factors together and feeds them into a Multilayer Perceptron (MLP) architecture. 
As such, the learned representations encode the non-linear interactions between different dimensions.

In addition to the aforementioned differences with RRN, our \model\ also differs from it with respect to the input to the LSTM module. 
To predict the next rating, the input to RRN's LSTM is the previous rating sequence with a fixed length, while the LSTM module of \model\ takes the representation vectors from previous time slots. 
Furthermore, \model\ is different from the recent work on neural collaborative filtering (NCF)~\cite{he2017neural} and collaborative deep learning (CDL)~\cite{wang2015collaborative}, whereas they infers the users' and items' latent embeddings under a static scenario. 
Additionally, different from CDL, \model\ does not need auxiliary information and domain knowledge to determine the effectiveness of features. 
The advantage of the \model\  model lies in its complete utilization of each dimension of the relational data. 


To sum up, the main contributions of this paper include:
\begin{itemize}[leftmargin=*]
\item To the best of our knowledge, we present the first model to generalize tensor factorization with deep neural networks, empowering it to model time-evolving multi-dimensional data.  We call this \model. 
\item We incorporate the multiple-layer perceptron architecture in \model\ for modeling the non-linearities in relational data, eliminating the linear limitation of dot product used in conventional tensor factorization. 
\item We perform extensive experiments for the problems of rating prediction in the Netflix dataset and link prediction in the Github dataset, demonstrating the significant improvements over state-of-the-art baselines, such as RRN and NCF. 
\end{itemize}

\section{Problem Formulation and \\Tensor Factorization (TF)}
\label{sec:model}

In this section, we present the notations and problem formulation. We also provide a quick review of the conventional Tensor Factorization (TF) model and discuss its limitations in modeling the temporal dynamics of relational data.

\subsection{Problem Formulation}
We consider a dynamic scenario wherein there exists evolving pair-wise relations between multiple types of entities (e.g., user, item, and time), such as Netflix users' ratings to various movies on different days, and Github users' repository forks during a month. 
To model the relationships among entities over time, we use a tensor to represent their time-evolving interactions. 

In this work, we focus on the three-way tensor with a temporal dimension. Formally, we construct a tensor $\mathcal{X}\in \mathbb{R}^{I\times J\times K}$ denoting the first-order tensor of each dimension with a size of $I$, $J$ and $K$, respectively. We denote the entry in the tensor $\mathcal{X}$ as $x_{i,j,k}$ to represent the interactions among different dimensions which are indexed by $i$, $j$ and $k$, respectively. For example, in relational rating data, $x_{i,j,k}$ can represent: (i) the quantitative rating score of $i$-th user on $j$-th item in $k$-th time slot, and (ii) the binary interactions (links) between $i$-th and $j$-th nodes at time slot $k$.

\textbf{Problem Formulation. } Based on the above definitions, we use $x_{i,j,k}$ to represent the interactions among three dimensions in tensor $\mathcal{X}$ which are observed from relational data. The objective of this work is to learn a predictive model that can effectively infer the unknown values in $\mathcal{X}$ with the observed ones.


\subsection{Conventional Tensor Factorization}
The key idea of tensor factorization is learning connections among the observed values in a tensor in order to infer the missing ones. The most common mechanism of tensor factorization is CANDECOMP/PARAFAC (CP)~\cite{carroll1970analysis}, which decomposes a tensor into multiple low-rank latent factor matrices representing each tensor-dimension. For example, movie rating dataset can be viewed as a tensor with $3$-dimensions: user, movie, and time. The latent factor matrices in this case, measure the latent factors of each dimension. The latent factors on user-dimension may be users' genre or rating preferences, the factor matrix on movie-dimension may model the movie plot, starring information, and many other features, the time-dimension factors could be specific time related information such as holiday season or special events. With these three matrices, the ratings can be obtained by a simple dot product across the matrices. Following the convention in Representation Learning (RL) literature, in this work we will refer inherent factor vectors as lower-dimensional embedding or representation vectors interchangeably~\cite{bengio2013representation}.


Formally, TF factorizes a tensor into three different matrices $U\in \mathbb{R}^{I\times L}$, $I\in \mathbb{R}^{J\times L}$ and $T\in \mathbb{R}^{K\times L}$, where $L$ is the number of latent factors (indexed by $l$). We define the factorization of tensor $\mathcal{X}$ as:
\begin{align}
\mathcal{X} \approx \sum_{l=1}^L U_{:,l} \circ I_{:,l} \circ T_{:,l},
\label{eq:tensor_decomp}
\end{align}
\noindent where $U_{:,l}$, $I_{:,l}$ and $T_{:,l}$ represents the $l$-th column of matrices $U$, $I$ and $T$ respectively. $\circ$ denotes the vector outer product. Each entry $x_{i,j,k}\in\mathcal{X}$ can be computed by the inner-product of three $L$-dimensional vectors as follows:
\begin{align}
\hat{x}_{i,j,k} \approx < U_i, I_j, T_k> \equiv \sum_{l=1}^L U_{i,l} I_{j,l} T_{k,l}.
\label{eq:X_i_j_k}
\end{align}
\noindent  
The objective of tensor factorization is to learn $U$, $I$ and $T$ using maximum likelihood estimation. We further define $U_i$, $I_j$ and $T_k$ to index the row of $U\in \mathbb{R}^{I\times L}$, $I\in \mathbb{R}^{J\times L}$ and $T\in \mathbb{R}^{K\times L}$, respectively. After the model learns $U$, $I$, and $T$ from the observed $\mathcal{X}$, one can easily fill in the missing values in $\mathcal{X}$ using Eq.~\ref{eq:X_i_j_k}.

However, two significant limitations exist in conventional tensor factorization model: (i) it fails to capture temporal dynamics because the interaction prediction $\hat{x}_{i,j,k}$ only depends on current timeslot $T_{k,:}$; (ii) it is a linear model which cannot deal with complex non-linear interactions that exist in real-world relational data. In the present work we aim to explicitly incorporate temporal dynamics into tensor factorization frameworks and model the non-linear interactions across different latent factors.

\section{\full\ Framework}
\label{sec:solution}

In this section, we present the \full\ (\model) framework, which is capable of learning implicit time-evolving interactions in relational data. We first introduce the general framework of \model\ to elaborate the motivation of the model and then present details of \model\ in the following subsections.
\subsection{General Framework}

Our \model\ is a multi-layer representation learning model which pursues a full neural treatment of tensor factorization to explicitly model the time-evolving interactions between different dimensions. We include the model architecture in Figure~\ref{fig:fra} and present the pseudo code of \model\ in Algorithm~\ref{alg:ntf}. The motivations of designing our model are listed as follows:
\begin{itemize}[leftmargin=*]
\item To address the data sparsity challenge, in our raw time embedding layer, we transform the element from the first-order tensor of temporal dimension with one-hot encoding and then project them into embedding space. In this way, we can address the sparse tensor challenge by using the latent vectors to represent elements of temporal dimension instead of using hand-crafted features.

\item To capture the complex temporal dynamics, with the generated embedding vectors as input, we utilize LSTM to encode the evolving interactions addressing the issues of long-term dependencies and vanishing gradients in recurrent neural networks~\cite{jozefowicz2015empirical}. Note that there are other variants of gated recurrent neural networks, such as Gated Recurrent Unit (GRU)~\cite{chung2015gated}. LSTM and GRU models bear some resemblance in the architecture and often provides similar results~\cite{chung2014empirical}. This work chooses LSTM as the encoder for the temporal dimension of the tensor because it is slightly more general compared to others~\cite{wu2017recurrent}. As a general tensor factorization framework, \model~is also flexible to integrate other variants of recurrent neural networks.

\begin{figure}[!t]
    \vspace{-0.2in}
    \begin{adjustbox}{max width=0.48\textwidth}
        \includegraphics[width=\linewidth]{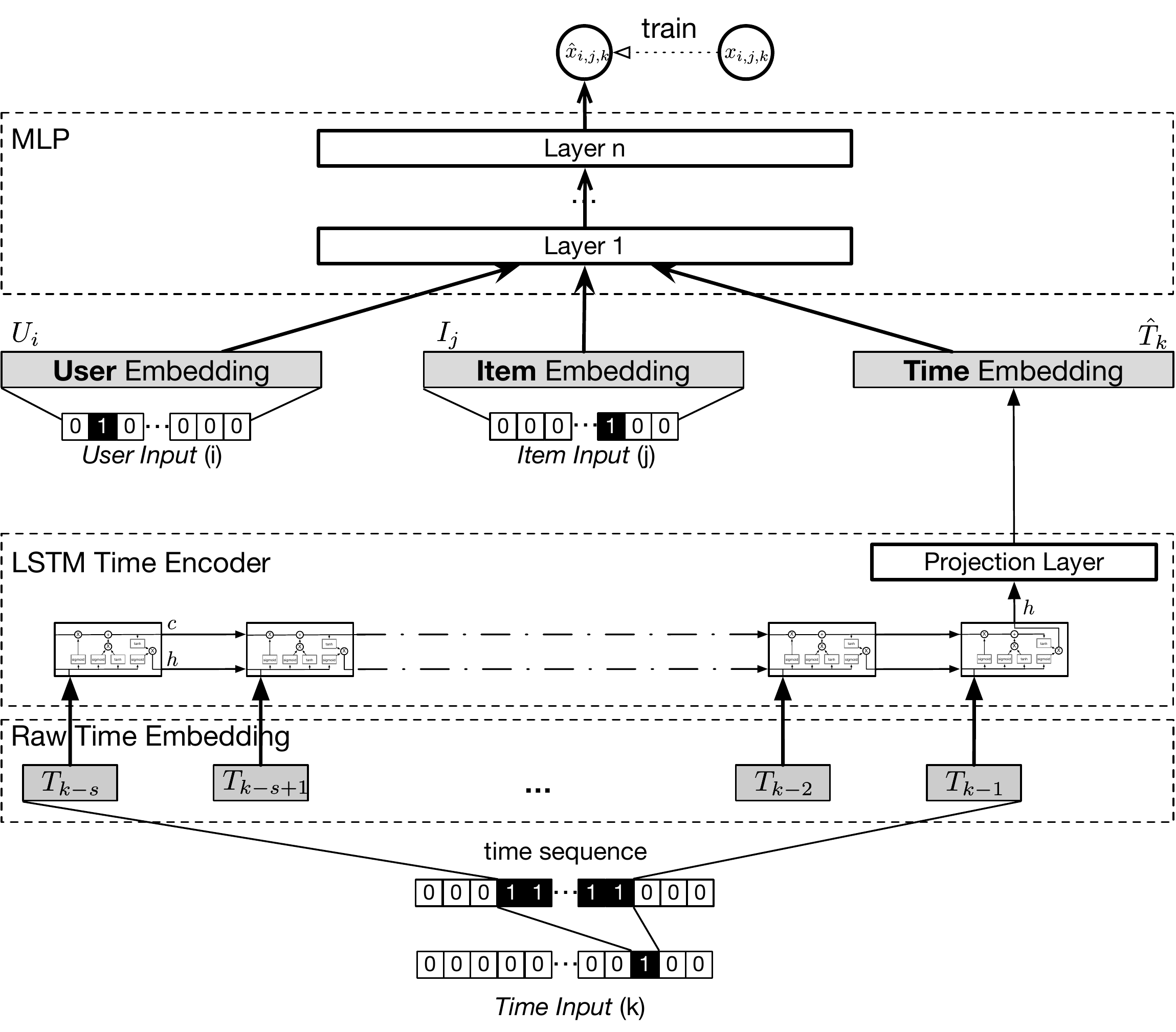}
    \end{adjustbox}
    \vspace{-0.5cm}
    \caption{The \full\ (\model) Framework. }
    \label{fig:fra}
    \vspace{-0.3in}
\end{figure}

\item To model the non-linearity of multi-dimensional interactions, we decide to use a Multi-layer Perceptron (MLP) on top of the first two layers. Let us consider the generated tensor with user-item-time dimensions as an example. The projected temporal embedding vectors will be fed into a multi-layer neural architecture together with the embedding vectors of users and items. This enables \model~to incorporate the learned complex dependencies in the temporal dimension into the factorization process as constrains. By doing so, we can detect the implicit patterns of user-item-time interactions through each layer in the MLP framework and model the non-linear combinations of latent factors to make better predictions. Finally, the latent vectors will be mapped to quantitative values (\ie, $\hat{x}_{i,j,k}$) which represents future interactions across different dimensions.
\end{itemize}

\begin{algorithm}[ht]
\small
    \SetKwData{User}{$\mathbf{U}$}
    \SetKwData{Item}{$\mathbf{I}$}
    \SetKwData{Time}{$\mathbf{T}$}
    \SetKwData{OneUser}{U}
    \SetKwData{OneItem}{I}
    \SetKwData{SeriesTime}{$\hat{\mathbf{T}}$}
    \SetKwData{Tindex}{t}
    
    \SetKwData{Tensor}{$\mathcal{X}$}
    \SetKwData{Train}{$\mathbf{X}$}
    \SetKwData{TrainBatch}{$\mathsf{T_{batch}}$}
    
    \SetKwData{Rtrue}{$x_{i,j,k}$}
    \SetKwData{Rpred}{$\hat{x}_{i,j,k}$}
    
    \SetKwData{Step}{$s$}
    \SetKwData{Batch}{$b_{\textsf{size}}$}
    \SetKwFunction{Sample}{sample}
    
    \SetKwFunction{LSTM}{LSTM}
    \SetKwData{Null}{\textsf{NULL}}
    \SetKwData{LSTMRes}{c}
    \SetKwFunction{MLP}{MLP}
    \SetKwData{OtherPara}{$\theta$}
    \SetKwFunction{Projection}{Projection}
    
    \SetKwData{Loss}{$\mathcal{L}$}
    
    \SetKwInOut{Parameter}{Paras}

    \KwIn{Tensor~\Tensor$\in \mathbb{R}^{I\times J\times K}$, observed interaction set~\Train, sequence length \Step, and batch size \Batch.}
    \Parameter{Embedding matrices~$\User\in\mathbb{R}^{I\times L}$, $\Item\in\mathbb{R}^{J\times L}$, $\Time\in\mathbb{R}^{K\times L}$, and other hidden parameters \OtherPara.}
    
    \emph{Initialize all parameters}\;
    \tcp{Sample a minibatch of size \Batch.}
    \ForEach{$\TrainBatch=\Sample(\Train,\Batch)$}{
        \ForEach{$\langle i,j,k\rangle \in\TrainBatch$} {
        \tcc{Gather embeddings for all dimensions.}
            $\OneUser_i=\User[i,:]$, $\OneItem_j=\Item[j,:]$\;
            
            $c=c_0, h=h_0$\tcp*{init. hidden states}
            \For{$\Tindex \gets (k - \Step)$ \KwTo $(k - 1)$}{
                $c,h = \LSTM(\Time[\Tindex,:], c, h)$\tcp*{according to Eq.\ref{eq:lstm_hidden}}
            }
            $\SeriesTime_k=\Projection(h)$\tcp*{encoded time embedding}
            
            $\Rpred=\MLP([\OneUser_i;\OneItem_j;\SeriesTime_k])$\tcp*{according to Eq.\ref{eq:mlp}}
            $\Rtrue=\Tensor[i,j,k]$\;
            update loss~\Loss w.r.t $\frac{1}{2}(\Rtrue-\Rpred)^2$\;
            
        }
    \emph{update all parameters w.r.t \Loss}\;

    }
    \caption{Training the NTF model.}\label{alg:ntf}
\end{algorithm}

\subsection{Modeling Temporal Dynamics via LSTM}
Recurrent Neural Network (RNN) has been widely used to address various challenges in time series analysis. However, two key limitations exist in the basic RNN. \emph{First}, it is difficult to train due to the problem of gradient vanishing and exploding, since the gradient might approach zero (vanish) or diverge (explode) as it is propagated
back through time steps during the training process. \emph{Second}, it is incapable to model long-distance dependencies in sequence data. To obviate the above problems and make the architecture more effective, LSTM is introduced as a special kind of RNN to model long-term dependencies and addresses the vanishing gradient problem by developing a more complicated hidden unit. In particular, LSTM proposes to derive the vector representations of hidden states $h_t$ and $c_t$ for each time step $t$ as follows:
\begin{align}
i_t=&\sigma(W_i h_{t-1}+V_i x_t+b_i) \nonumber\\
o_t=&\sigma(W_o h_ {t-1}+V_o x_t+b_o) \nonumber\\
f_t=&\sigma(W_f h_{t-1}+V_f x_t+b_f) \nonumber\\
\widetilde{c_t} = &\phi(W_c h_{t-1}+V_c x_t+b_c) \nonumber\\
c_t = &f_t \odot c_{t-1} + i_t \odot \widetilde{c_t} \nonumber\\
h_t = &o_t \odot \phi(c_t)
\label{eq:lstm_hidden}
\end{align}

\noindent where $W_*\in \mathbb{R}^{d_s\times d_s}$ represents the transformation matrix from the previous state (\ie, $c_{t-1}$ and $h_{t-1}$) to LSTM cell and $V_*\in \mathbb{R}^{d_x\times d_s}$ are the transformation matrices from input to LSTM cell, where $d_x$ and $d_s$ denotes the dimension of input vectors and hidden states, respectively. Furthermore, $b_*\in \mathbb{R}^{d_s}$ is defined as a vector of bias term. $\sigma (\cdot)$ and $\phi (\cdot)$ represents the sigmoid and tanh function, respectively. The $\odot$ operator denotes the element-wise product. In Eq.~\ref{eq:lstm_hidden}, $i_t$, $o_t$, and $f_t$ represents input gate, output gate and forget gate, respectively. For simplicity, we denote Eq.~\ref{eq:lstm_hidden} as $[c_t, h_t] = \text{LSTM}(*, c_{t-1}, h_{t-1})$ in the following subsections.


\subsection{Fusion of LSTM and TF}
In this subsection, we present how we fuse LSTM and TF under the \model\ framework to model the time-evolving interactions across three dimensions $\mathbb{R}^{I}$, $\mathbb{R}^{J}$ and $\mathbb{R}^{K}$. In relational data, dynamic characteristics are observed across different time slots. In this work, we consider temporal factors affecting the interactions over time based on a global trend by assuming that the interactions between multi-dimensions evolve in a smooth way. In particular, to capture the temporal smoothness, we further assume that the embedding vectors of the temporal dimension depends on embedding vectors from previous $s$ time slots. In \model, we predict the embedding vector in current time slot based on the embedding vectors from past $s$ time slots using LSTM.

To encode the evolving temporal hidden factors, our LSTM encoder generates embedding without using hand-crafted features (\eg, the day of a week). We formally define the hidden states $c_t$ and $h_t$ in encoding the contextual sequence as:
\begin{align}
[c_1, h_1] =& \text{LSTM}(T_{k-s}, c_0, h_0)\nonumber\\
&\cdots \nonumber\\
[c_{s-1}, h_{s-1}] =& \text{LSTM}(T_{k-1}, c_{s-2}, h_{s-2})
\label{eq: temporal embedding}
\end{align}
\noindent Using the last hidden state vector $h_{s-1}$, we can define the embedding vector $\hat{T}_k$ through Projection Layer as:
\begin{align}
\hat{T}_k =&\phi_n (W_T h_{s-1} +b_T)
\label{eq: temporal embedding}
\end{align}
where $W_T\in \mathbb{R}^{s\times L}$ is the projection matrix, $b\in \mathbb{R}^{L}$ is the projection bias. $\phi$ are activation functions that we define later. 

Finally, the predictive value $\hat{x}_{i,j,k}$ could be derived by the dot product of $U_i$, $I_j$, $\hat{T}_k$ according to the conventional tensor factorization. With the increasing utilization of deep neural networks to handle complicated non-linearity in image and text data~\cite{collobert2011natural,rigamonti2011sparse}, it is intuitive to explore the non-linear interactions in relational data. To address this issue, we concatenate embedding vectors of $U_i$, $I_j$, $\hat{T}_k$ together and consider them as input to the Multi-Layer Perceptron (MLP) and output $\hat{x}_{i,j,k}$. In this way, we can address the limitation caused by the dot product in tensor factorization (as introduced in Section~\ref{sec:model}) with a neural network architecture to capture non-linear interactions by concatenating latent factors from the previous embedding layer. Formally, we present MLP as:
\begin{align}
Z_1=&\phi_1 (W_1[U_i; I_j; \hat{T}_k] +b_1) \nonumber\\
&\cdots \nonumber\\
Z_n=&\phi_n (W_L Z_{n-1} +b_L)\nonumber\\
\hat{x}_{i,j,k} =& W_o Z_n +b_o
\label{eq:mlp}
\end{align}
\noindent where $n$ represents the number of hidden layers which is indexed by $l$ and $;$ represents the concatenate operation. For the $Z_l$ layer, $\phi_l$, $W_l$ and $b_l$ represent the activation function (e.g., $ReLU$ or $tanh$ function) of MLP layers, weight matrix and bias vector, respectively. We further specify the activation function as~\emph{sigmoid} (denoted as $\sigma$) to output the quantitative values representing multi-dimensional interactions. In the experiments, we investigate the effect of number of layers in MLP.

\subsection{Learning Process}
In this subsection, we first describe the learning process of our \model\ framework. Then, we further utilize the advanced technique, \ie, \emph{batch normalization}, to optimize the \model.

\subsubsection{The Objective Function.}
As we introduced in the Section~\ref{sec:model}, our objective is to derive the value of $\hat{x}_{i,j,k}$ which denotes the interactions between $i$-th, $j$-th and $k$-th elements of first-order tensor $\mathbb{R}^{I}$, $\mathbb{R}^{J}$ and $\mathbb{R}^{K}$ in $\mathcal{X}$, respectively. We formally define our objective function in factorization procedure as follows:
\begin{align}
\argmin_{U,I,T} \frac{1}{2} \sum_{(i,j,k)\in \mathbf{X}}  (x_{i,j,k}-\hat{x}_{i,j,k})^2
\label{eq:loss_fun}
\end{align}
\noindent where $\mathbf{X}$ denotes the set of observed interactions in tensor $\mathcal{X}$. The \model\ can be learned by 
minimizing the above loss function between the observed interaction data and the factorization representation. The above optimization problem can be efficiently solved using a popular optimizer Adaptive Moment Estimation (Adam). The reasons to choose Adam are mainly two-folds: (i) it can automatically tune the learning rate during the training, and (ii) it often provides faster convergence compared with stochastic gradient descent algorithm.

\subsubsection{Batch Normalization.}
In the training process of neuron network models, their performances could be degraded by covariance shift~\cite{shimodaira2000improving}.
To tackle this challenge, \emph{Batch Normalization (BN)} has been proposed to normalize the input data from previous layer before sending it to the next layer as input~\cite{ioffe2015batch}. In \model\ framework, we apply \emph{BN} to reduce the internal covariance shift by transforming the input to zero mean/unit variance distributions in each mini-batch training. In \model, we apply the BN to LSTM to avoid the deceleration in training process as:
\begin{align}
\widetilde{c_t} = &\phi(BN(W_c h_{t-1}+V_c x_t+b_c))
\end{align}
\noindent where $BN(\cdot)$ stands for the batch normalization operation. \emph{BN} is also applied to Projection Layer.




\section{Evaluation}
\label{sec:eval}

We demonstrate the effectiveness of \model\ with two real-world applications on dynamic relational data, \ie, regression-\emph{rating prediction} and classification-\emph{link prediction} corresponding to predicting quantitative (rating scores) and binary (existence of links) interactions, respectively. In particular, we aim to answer the following questions:
\begin{itemize}[leftmargin=*]
    \item \textbf{Q1}: How does our \model\ framework perform as compared to the state-of-the-art techniques in rating prediction task?\\\vspace{-0.1in}
    \item \textbf{Q2}: How does our \model\ framework work for link prediction task when competing with baselines?\\\vspace{-0.1in}
    \item \textbf{Q3:} Does \model\ consistently outperform other baselines in terms of prediction accuracy with respect to different time windows with different training and testing time period?\\\vspace{-0.1in}
    \item \textbf{Q4}: How is the performance of \model\ variants with different combinations of key components in the joint framework?\\\vspace{-0.1in}
    \item \textbf{Q5}: How different hyper-parameter settings (e.g., embedding size and number of hidden layers) affect the performance of \model?
\end{itemize}

\subsection{Experimental Setup}
\subsubsection{\bf Data}
In our evaluation, we perform experiments on two types of dynamic relational datasets and corresponding tasks, namely: (i) rating prediction on Netflix movie rating data; (ii) link prediction on Github archive data. \\\vspace{-0.1in}

\noindent \textbf{Netflix Rating Data}. This movie rating dataset, which was collected between Jan 2002 and Dec 2005, has been widely used in rating prediction evaluation~\cite{xiong2010temporal,shih2016dynamically}. In the Netflix dataset, users rate a movie using a 1 (worst) to 5 (best) scale, the given score is also associated with a rating date to denote when the rating was reported. We generate tensor $\mathcal{X}$ by associating each movie with the users who rated this movie on different months. In particular, if $i$-th user rated $j$-th movie on $k$-th month (the time slots we used for evaluation correspond to calendar months) in the dataset, the element $x_{i,j,k}$ is in the interaction set $\mathcal{X}$.\\\vspace{-0.1in}

\noindent \textbf{Github Archive Data}. This dataset was collected from Github to record the \emph{fork} actions of users on repositories. Specifically, forking a repository allows users to freely experiment with changes without affecting the original project. Note that for any repository a user can only fork it once. The collection lasts for $6$-month (Jan 2017 to Jun 2017) and the time information is provided. In this dataset, an edge (i,j,k) is generated when $i$-th user forks the $j$-th repository at time slot $k$ (the time slots we used for evaluation correspond to calendar weeks). We set the element $x_{i,j,k}$ in $\mathcal{X}$ tensor to 1 if edge $(i,j,k)$ exists in the dataset and 0 otherwise.



\begin{table}[t!]
\centering
\footnotesize
\caption{The Statistics of Datasets}
\vspace{-0.1in}
\begin{tabular}{l| c| c| c| c}
\toprule
& \# of Users & \# of Items & \# of Ratings & Time Span\\
\cmidrule(l){2-5}
Netflix  & 68,079 & 2,328& 12,326,319  & 36 months\\
\midrule
& \# of Users & \# of Projects & \# of Fork & Time Span\\
\cmidrule(l){2-5}
Github & 81,001 & 72,420 & 1,396,115  & 21 weeks \\
\bottomrule
\end{tabular}
\label{tab:data}
\vspace{-0.2in}
\end{table}

Table~\ref{tab:data} summarizes the statistics of the above two datasets. To better understand the effectiveness of \model\ in modeling temporal dynamics, we evaluate \model\ on different time windows with different training and testing period. Additionally, to evaluate the ability of our \model\ to model temporal dynamics with time-evolving interactions among data, we remove users and items in the testing data which are not included in the training data. Table~\ref{tab:data_detail} shows the details to summarize of the experimental settings by varying time windows.


\begin{table}[thb!]
\centering
\footnotesize
\vspace{-0.1in}
\caption{Different Netflix data splits used in rating prediction. }
\vspace{-0.1in}
\begin{tabular}{p{0.53cm} | p{0.4cm} | p{0.8cm} | p{0.75cm} | p{0.65cm} | p{2cm}| p{1cm}}
\toprule
\# of users & \# of items & training size & validation size & testing size & training period & testing period\\
\midrule
28,077 & 1,772 & 1,454,868 & 145,486 & 225,787 & Jan 2003-Dec 2004 & Jan 2004 \\
\midrule
32,637 & 1,862 & 1,910,235 & 191,023 & 247,956 & Mar 2003-Feb 2004 & Mar 2004 \\
\midrule
37,060 & 1,937 & 2,437,882 & 243,788 & 270,312 & May 2003-Apr 2004 & May 2004 \\
\midrule
40,922 & 1,986 & 2,941,755 & 294,175 & 288,037 & Jul 2003-Jun 2004 & Jul 2004 \\
\midrule
44,072 & 2,055 & 3,388,774 & 338,877 & 305,693 & Sep 2003-Aug 2004 & Sep 2004 \\
\midrule
47,480 & 2,126 & 3,869,204 & 386,920 & 357,878 & Nov 2003-Oct 2004 & Nov 2004 \\
\bottomrule
\end{tabular}
\label{tab:data_detail}
\vspace{-0.2in}
\end{table}

\subsubsection{\bf Baselines}
Because rating prediction and link prediction are two different tasks and have different representative baselines. Here, we summarize the compared baselines of these two tasks separately. In addition, the reason to compare \model\ with matrix factorization methods rather than tensor factorization schemes is mainly twofold: (1) it is difficult to apply tensor factorization with temporal dimension to make predictions due to the ignorance of temporal dependencies between time slots. (2) There is no duplicate interactions (\ie, ratings and links) existed between users and items in different time slots in both Netflix and Github datasets.

\noindent \textbf{Rating Prediction and Inference}. For the rating prediction, we consider three types of baselines: representative matrix factorization for recommendation systems, neural network based collaborative filtering methods for recommendations or predictive analytics, and variants of Recurrent Neural Network models for time series prediction.
\begin{itemize}[leftmargin=*]
    \item \textbf{Probabilistic Matrix Factorization (PMF)}~\cite{mnih2008probabilistic}: it is a probabilistic method for matrix factorization, which assigns a D-dimensional latent feature vector (following Gaussian distributions) for each user and item. The ratings are derived from the inner-product of corresponding latent features.
    \item \textbf{Bayesian Probabilistic Matrix Factorization (BPMF)}~\cite{salakhutdinov2008bayesian}: extended from the PMF, BPMF learns the latent feature vector for each user and item by Monte Carlo Markov Chain method, which is able to address the overfitting issue. 
    \item \textbf{Bayesian Probabilistic Tensor Factorization (BPTF)}~ \cite{xiong2010temporal}: it is a bayesian probabilistic tensor factorization method for modeling evolving relational data.
    \item \textbf{Temporal Deep Semantic Structured Model (TDSSM)}~\cite{song2016multi}: this method is a temporal recommendation model which combines traditional feedforward networks (DSSM) with LSTM, to capture temporal dynamics of users' interests.
    \item \textbf{Recurrent Recommendation Networks (RRN)}~\cite{wu2017recurrent}: it aims to predict future interactions between users and items by specifying two embedding vectors (stationary and dynamic) for both user and item. The dynamic embedding vectors are inferred with LSTM model based on historical ratings.
    \item \textbf{Neural Collaborative Filtering (NCF)}~\cite{he2017neural}: it proposed a framework for collaborative filtering based on neural network architecture to model the interactions between users and items.
\end{itemize}

\subsubsection{\bf Evaluation Protocols}
In our evaluation, we split the datasets into training, validation and test sets. We use the validation datasets to tune hyper-parameters and test datasets to evaluate the final performance of all compared algorithms.
\begin{itemize}[leftmargin=*]
\item \textbf{Rating Prediction}. To evaluate the performance of all compared algorithms in predicting quantitative rating scores, we use~\emph{Root Mean Square Error (RMSE)} and~\emph{Mean Absolute Error (MAE)} which have been widely adopted in quantitative prediction tasks~\cite{fm2017neural}. Note that a lower RMSE and MAE score indicates better performance. The mathematical definitions of those metrics are presented as follows: $RMSE=\sqrt{\frac{1}{N} \sum_{(i,j,k)\in \mathbf{X}}   (x_{i,j,k} - \hat{x}_{i,j,k})^2}$, $MAE=\frac{1}{N} \sum_{(i,j,k)\in \mathbf{X}}  |x_{i,j,k} - \hat{x}_{i,j,k}|$, where $N$ denotes the number of observed elements in tensor  $\mathcal{X}$, $x_{i,j,k}$ and $\hat{x}_{i,j,k}$ represents the actual rating score and estimated rating score, respectively.
\item \textbf{Link Prediction}. To validate the performance of each method in predicting the existences of links, \emph{Precision}, \emph{Recall}, \emph{F1-score} and \emph{AUC} are used as evaluation metrics~\cite{scellato2011exploiting}.
\end{itemize}

\subsubsection{\bf Reproducibility}
We summarize the parameter settings of \model\ and experiments in Table~\ref{tab:para_setting}. In addition, we vary each of key parameters in \model\ and fix others to examine the parameter sensitivity. We implemented our framework based on TensorFlow and chose Adam~\cite{kingma2014adam} as our optimizer to learn the model parameters\footnote{Code of our model and baselines will be publicly available upon publication}. For all neural network baselines (\ie, NCF, RRN and TDSSM), we use the same parameters listed in Table~\ref{tab:para_setting}.

\begin{table}[thb!]
 \centering
 \small
 \vspace{-0.10in}
 \caption{Parameter Settings}
 \vspace{-0.1in}
 \begin{tabular}{l| c | l | c}
 \toprule
Parameter & Value & Parameter & Value \\
\midrule  
Hidden State Dimension & 32 & Embedding size & 32\\
\# of Time Steps & 5 & \# of Hidden Layers & 6\\
BN scale parameter &  0.99 & BN shift parameter &  0.001\\
Batch size & 256 & Learning rate & 0.001\\
\midrule 
\end{tabular}
 \vspace{-0.1in}
\label{tab:para_setting}
\end{table}


\begin{table}[t!]
\centering
\footnotesize
\vspace{-0.1in}
\caption{Baseline Summary}
\vspace{-0.1in}
\begin{adjustbox}{max width=\linewidth}
\begin{tabular}{p{1.7cm} | p{0.3cm} |p{0.5cm} | p{0.2cm} |p{0.2cm} | p{0.6cm} | p{0.4cm} |p{0.3cm} | p{0.4cm}| p{0.2cm} }
\toprule
Method & PMF & BPMF & AA & AP & TDSSM & RRN & NCF & BPTF & \model\\
\midrule
Rating Prediction & \checkmark & \checkmark& & & \checkmark& \checkmark &  \checkmark & & \checkmark\\
\midrule
Rating Inference & \checkmark & \checkmark& & & \checkmark& \checkmark &  \checkmark & \checkmark & \checkmark\\
\midrule
Link Prediction & \checkmark& \checkmark&  \checkmark & \checkmark & & & \checkmark& & \checkmark \\
\bottomrule
\end{tabular}
\end{adjustbox}
\label{tab:basaelines}
\vspace{-0.1in}
\end{table}

\noindent \textbf{Link Prediction}. 
In addition to the above \emph{NCF}, \emph{BPMF} and \emph{PMF} algorithms which have been applied to solve link prediction problem, we consider other two traditional link prediction baselines to compare the performance of \model\ in predicting future binary interactions (\ie, links) among users.
\begin{itemize}[leftmargin=*]
    \item \textbf{Preferential Attachment (PA)}~\cite{liben2007link}: PA assumes new connections are more likely to form between well-connected nodes.
    \item \textbf{Adamic Adar (AA)}~\cite{adamic2003friends}: AA smoothes the common neighbor method using neighbors' node degree. 
    \item \textbf{PMF}, \textbf{BPMF}, and \textbf{NCF}: as introduced above.
    \item  \textbf{TDSSM} and \textbf{RRN}: both methods require the historical sequences of items and users. However, the sequences can not be generated from the Github dataset due to the following two reasons: First, the unobserved interactions may be either negative or positive cases; Second, if we take all unobserved cases as negative, it is also difficult to locate them to specific time slots. 
\end{itemize}
For all embedding based and recurrent neural network based baselines, we use the same parameters as \model\ which are listed in Table~\ref{tab:para_setting}.

\subsubsection{\bf Variants of \model}

In addition to comparing \model\ with existing approaches, we are also interested in discovering the best way to model non-linear multi-dimensional interactions among different embeddings in the proposed \model\ framework. Namely, we aim to answer the following two questions: (1) does the selection of activation functions affect the performance of \model? and (2) is multiple-layer perceptron, helpful for learning non-linear interactions from multi-dimensional relational data? Hence, in the evaluation of \model\ framework, we consider four variants of \model: \emph{\model $dot$}: a simplified version of \model\ which does not use MLP to explore the non-linear interactions in relational data. Instead, it uses dot product to predict value $\hat{x}_{i,j,k}$, which is also applied in compared baselines. \emph{\model($ReLU$)}, \emph{\model($sigmoid$)}, and \emph{\model ($tanh$)}: the full version of \model\ that use different activation functions.

\begin{table*}[t!]
\centering
\renewcommand\arraystretch{1.2}
\vspace{-0.1in}
\caption{Performance of rating prediction for all compared algorithms \wrt. different time windows.}
\vspace{-0.1in}
\begin{adjustbox}{max width=.95\linewidth}
\begin{tabular}{l| c c | c c | c c |  c c |  c c |  c c} 
\toprule
Month
& \multicolumn{2}{c|}{2004-Jan}
& \multicolumn{2}{c|}{2004-Mar}
& \multicolumn{2}{c|}{2004-May}
& \multicolumn{2}{c|}{2004-Jul}
& \multicolumn{2}{c|}{2004-Sep}
& \multicolumn{2}{c}{2004-Nov}\\
\midrule
Metrics & RMSE & MAE & RMSE & MAE & RMSE & MAE & RMSE & MAE & RMSE & MAE & RMSE & MAE\\
\midrule
PMF                & $0.9385$ & $0.7331$ & $0.9274$ & $0.7263$ & $0.9243$ & $0.7171$ & $0.8968$ & $0.6972$ & $0.8978$ & $0.6923$ & $0.8885$ & $0.6856$ \\
BPMF               & $0.9879$ & $0.7686$ & $0.9829$ & $0.7659$ & $0.9741$ & $0.7541$ & $0.9449$ & $0.7319$ & $0.9415$ & $0.7264$ & $0.9312$ & $0.7165$\\
TDSSM              & $1.0031$ & $0.8001$ & $1.0386$ & $0.8488$ & $0.9897$ & $0.7886$ & $0.9365$ & $0.7442$ & $0.9259$ & $0.7306$ & $0.9401$ & $0.7413$\\
RRN                & $1.0062$ & $0.7936$ & $0.9901$ & $0.7798$ & $0.9721$ & $0.7584$ & $0.9328$ & $0.7276$ & $0.9574$ & $0.7529$ & $0.9227$ & $0.7146$\\
NCF                & $0.9498$ & $0.7517$ & $0.9364$ & $0.7357$ & $0.9421$ & $0.7408$ & $0.9069$ & $0.7148$ & $0.9080$ & $0.7071$ & $0.9089$ & $0.7068$\\
\midrule
\model $dot$           & $0.9869$ & $0.7763$ & $0.9736$ & $0.7702$ & $0.9600$ & $0.7523$ & $0.9231$ & $0.7221$ & $0.9237$ & $0.7186$ & $0.9123$ & $0.7089$\\
\model($ReLU$)        & $0.9192$ & $0.7204$ & $0.9111$ & $0.7169$ & $0.9127$ & $0.7131$ & $0.8823$ & $0.6869$ & $0.8871$ & $0.6923$ & $0.8753$ & $0.6787$\\
\model($sigmoid$)     & $0.9158$ & $0.7178$ & $0.9113$ &$0.7148$ & $0.9141$ & $0.7110$ & $0.8802$ & $0.6889$ & $0.8882$ & $0.6875$ & $0.8751$ & $0.6793$\\
\model($tanh$)        &$0.9178$ &$0.7187$ & $0.9128$ & $0.7185$ & $0.9135$ &$0.7111$ &$0.8834$ &$0.6896$ &$0.8865$ &$0.6941$ &$0.8779$ &$0.6783$\\
\bottomrule
\end{tabular}
\end{adjustbox}
\label{tab:esti_rating_time}
\end{table*}

\begin{table*}[t!]
\centering
\renewcommand\arraystretch{1.2}
\caption{Performance of rate prediction for all compared algorithms \wrt. different percentages of training data.}
\vspace{-0.1in}
\begin{adjustbox}{max width=.95\linewidth}
\begin{tabular}{l| c c | c c | c c |  c c | c c | c c} 
\toprule
Training ratio
& \multicolumn{2}{c|}{30\%}
& \multicolumn{2}{c|}{40\%}
& \multicolumn{2}{c|}{50\%}
& \multicolumn{2}{c|}{60\%}
& \multicolumn{2}{c|}{70\%}
& \multicolumn{2}{c}{80\%}\\
\midrule
Metrics & RMSE & MAE & RMSE & MAE & RMSE & MAE & RMSE & MAE & RMSE & MAE & RMSE & MAE\\
\midrule
PMF                & $0.9241$ & $0.7154$ & $0.9041$ & $0.6974$ & $0.8852$ & $0.6813$ & $0.8690$ & $0.6691$ & $0.8562$ & $0.6588$ & $0.8453$ & $0.6502$\\
BPMF               & $0.9328$ & $0.7204$ & $0.9103$ & $0.7049$ & $0.8920$ & $0.6904$ & $0.8781$ & $0.6778$ & $0.8673$ & $0.6698$ & $0.8578$ & $0.6616$\\
BPTF               & $0.9167$ & $0.7384$ &  $0.8964$ & $0.7032$ & $0.8896$ & $0.6971$ &  $0.8767$ & $0.6870$ & $0.8625$ & $0.6750$ & $0.8572$ & $0.6706$\\
TDSSM              & $0.9591$ & $0.7631$ & $0.9346$ & $0.7443$ & $0.9443$ & $0.7613$ & $0.8895$ & $0.6981$ & $0.8592$ & $0.6665$ & $0.8601$ & $0.6686$\\
RRN                & $0.9223$ & $0.7210$ & $0.9003$ & $0.7039$ & $0.8842$ & $0.6891$ & $0.8775$ & $0.6815$ & $0.8754$ & $0.6805$ & $0.8592$ & $0.6673$\\
NCF                & $0.9124$ & $0.7211$ & $0.8947$ & $0.7012$ & $0.8810$ & $0.6924$ & $0.8730$ & $0.6829$ & $0.8629$ & $0.6692$ & $0.8563$ & $0.6651$\\
\midrule
\model $dot$            & $0.9161$ & $0.7184$ & $0.8921$ & $0.6968$ & $0.8779$ & $0.6846$ & $0.8644$ & $0.6723$ & $0.8547$ & $0.6639$ & $0.8451$ & $0.6563$\\
\model($ReLU$)  & $0.8759$ & $0.6829$ & $0.8698$ & $0.6784$ & $0.8632$ & $0.6708$ & $0.8571$ & $0.6675$ & $0.8494$ & $0.6586$ & $0.8437$ & $0.6545$ \\
\model($sigmoid$)  & $0.8747$ & $0.6830$ & $0.8712$ & $0.6796$ & $0.8637$ & $0.6729$ & $0.8579$ & $0.6664$ & $0.8476$ & $0.6538$ & $0.8448$ & $0.6562$\\
\model($tanh$)  & $0.8781$ & $0.6852$ & $0.8726$ & $0.6819$ & $0.8661$ & $0.6743$ & $0.8666$ & $0.6741$ & $0.8560$ & $0.6649 $ & $0.8498$ & $0.6601$\\
\bottomrule
\end{tabular}
\end{adjustbox}
\label{tab:esti_rating_train}
\end{table*}

\begin{figure*}[thb!]
    \centering
    \vspace{-0.1 in}
    \subfigure[][Embedding Size]{
        \centering
        \includegraphics[width=0.22\textwidth]{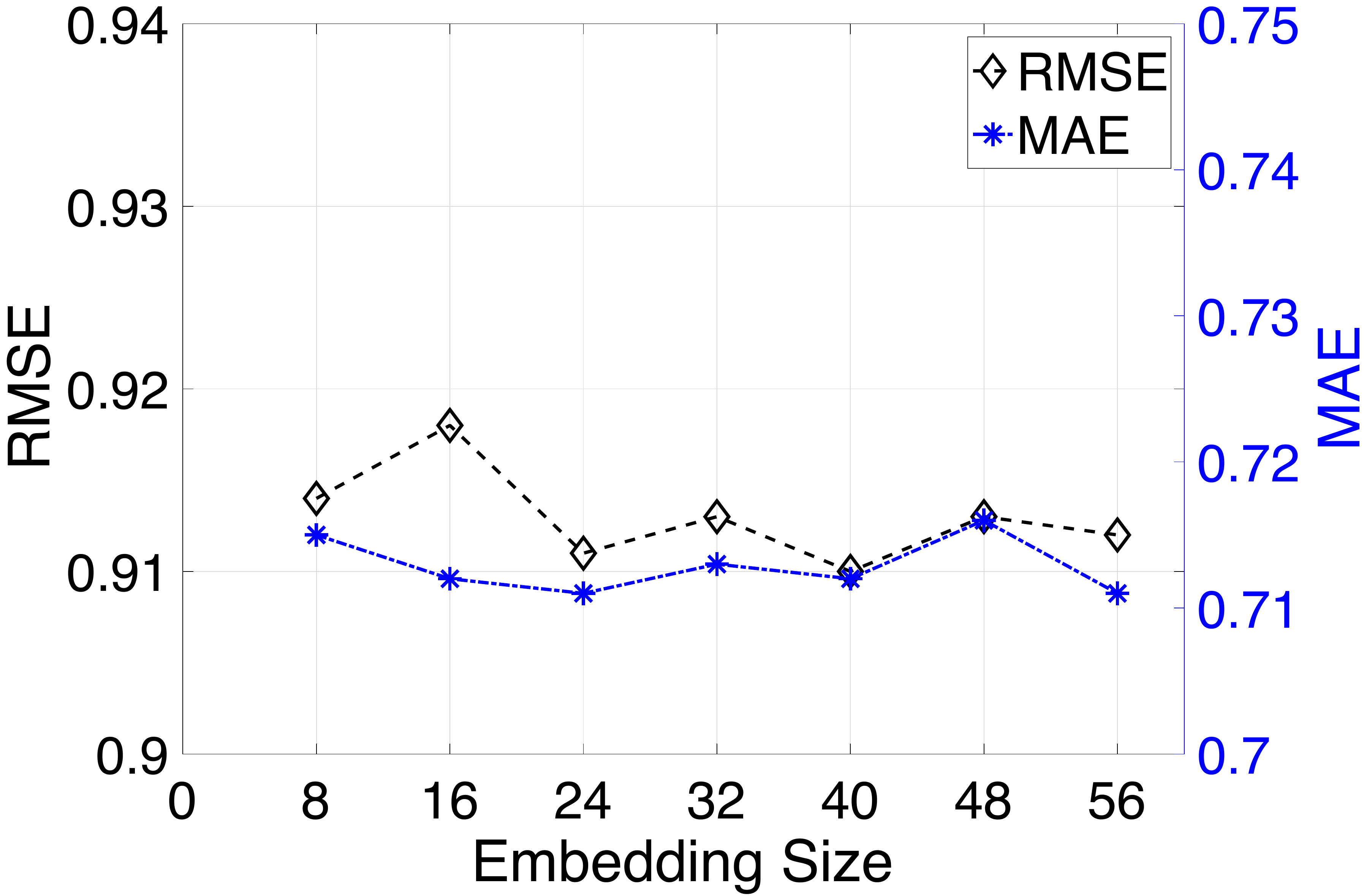}
        \label{fig:pre_para_embedding}
        }
    \subfigure[][\# of Hidden Layers]{
        \centering
        \includegraphics[width=0.22\textwidth]{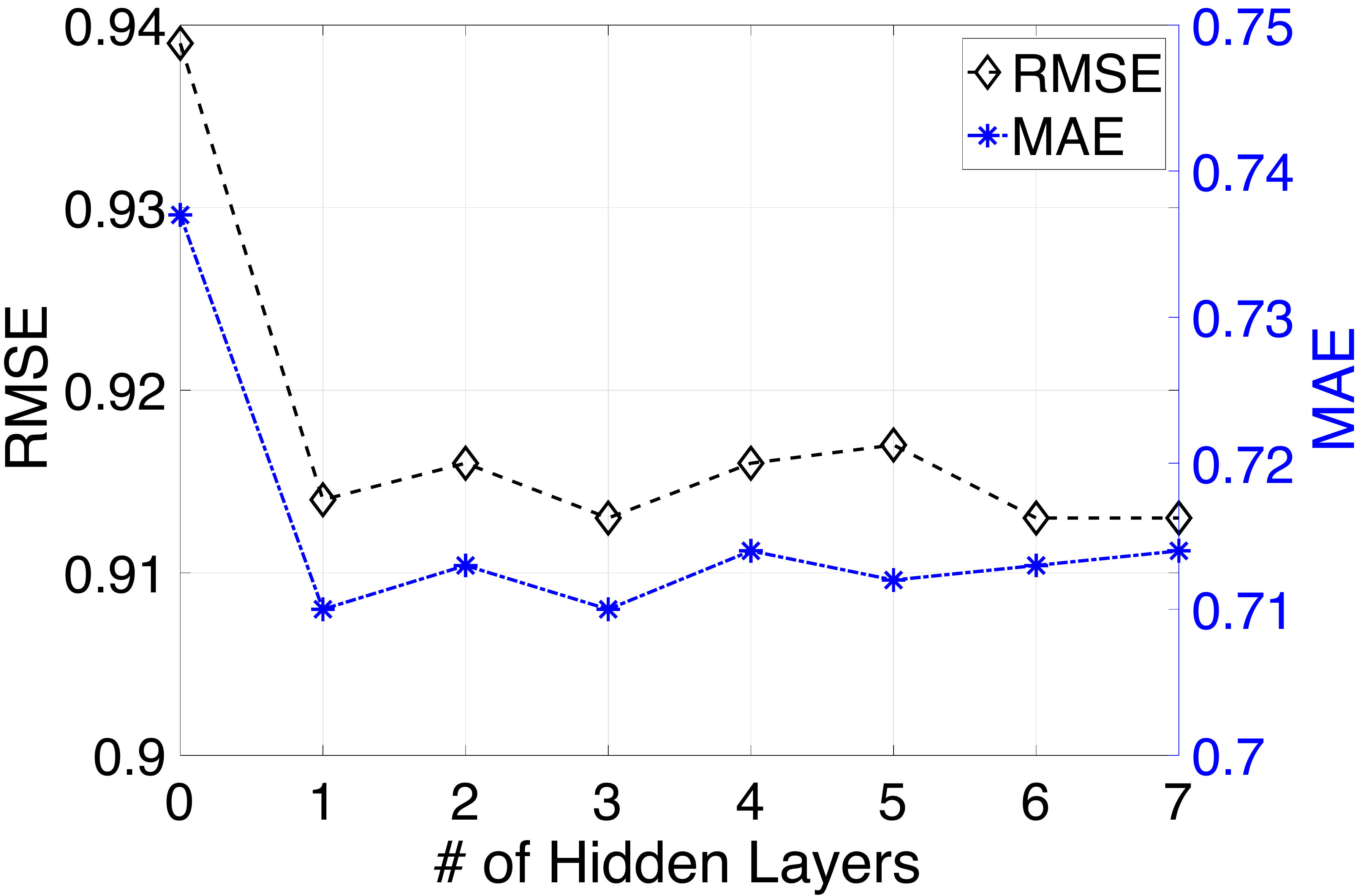}
        \label{fig:pre_para_layer}
        }
    \subfigure[][\# of Time Steps]{
        \centering
        \includegraphics[width=0.22\textwidth]{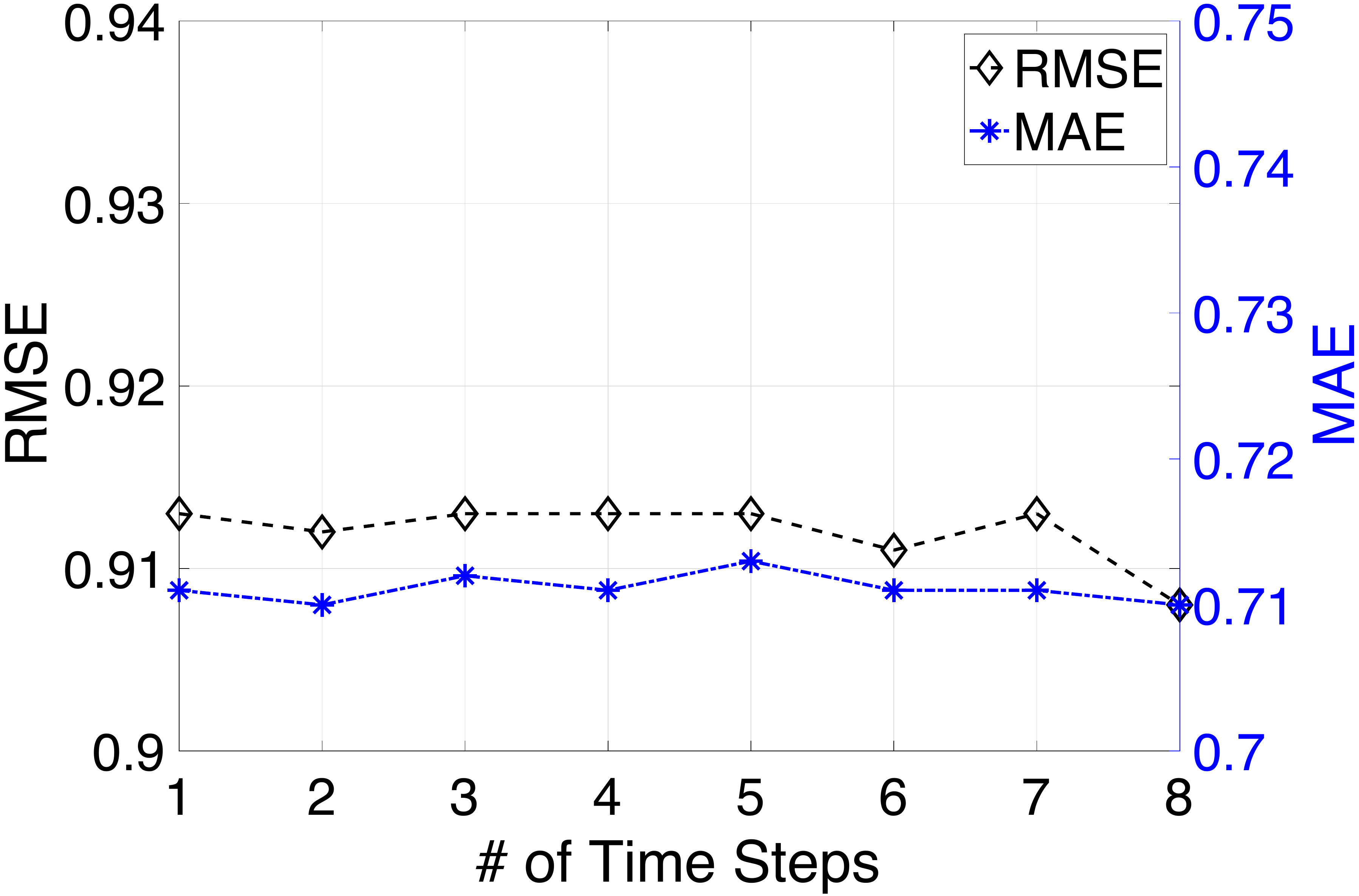}
        \label{fig:pre_para_sequence}
        }
    \subfigure[][Hidden State Dimension]{
        \centering
        \includegraphics[width=0.22\textwidth]{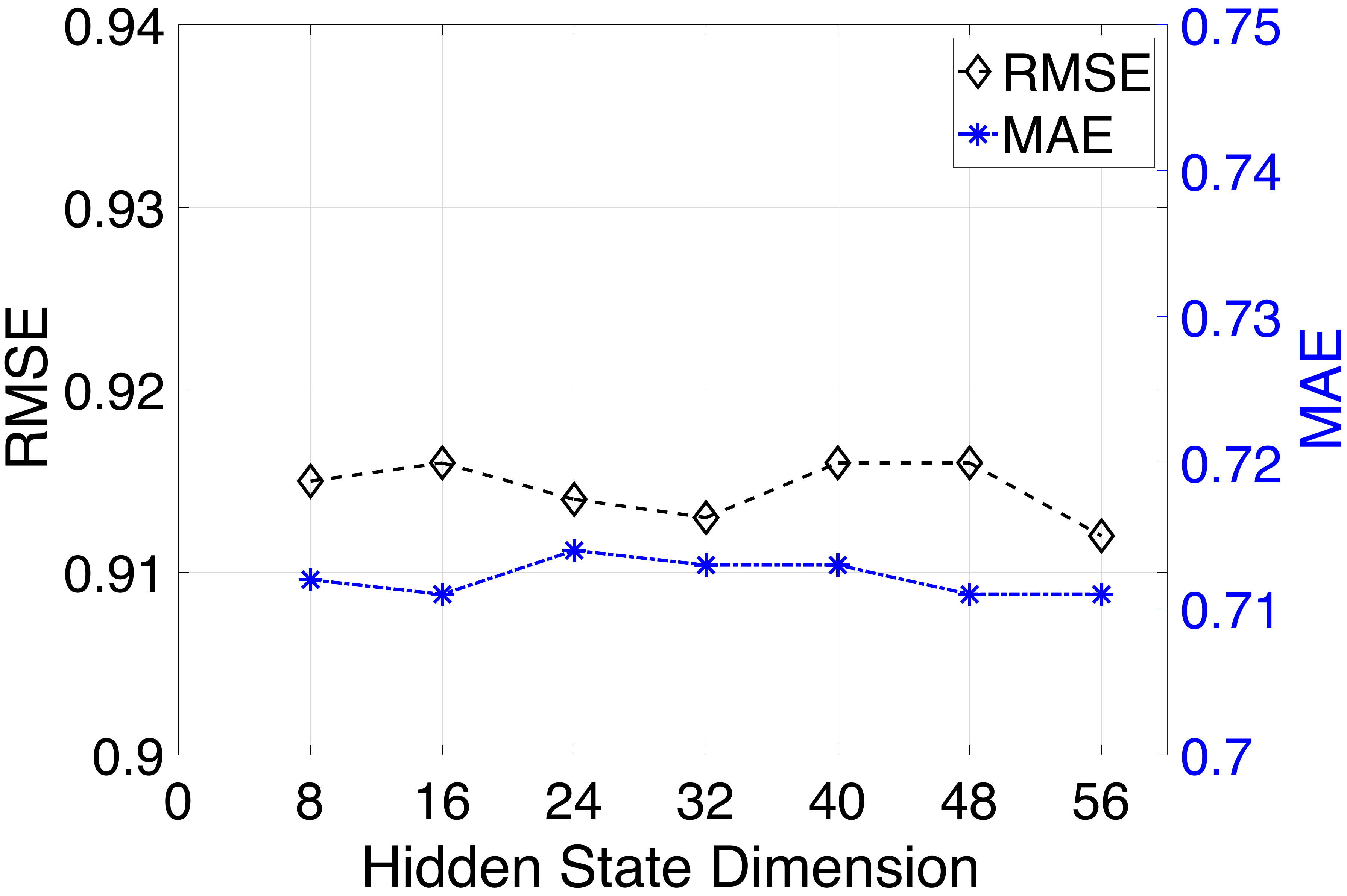}
        \label{fig:pre_para_state}
        }
    \vspace{-0.1in}
    \caption{Parameter sensitivity of \model($ReLU$) in rating prediction (data from May 2003 to Apr 2004 is used for training and validation, data from May 2004 is used for testing).}
    \label{fig:para_sen_rating_pre}
\end{figure*}

\begin{figure*}[thb!]
    \centering
    \vspace{-0.1 in}
    \subfigure[][Embedding Size]{
        \centering
        \includegraphics[width=0.22\textwidth]{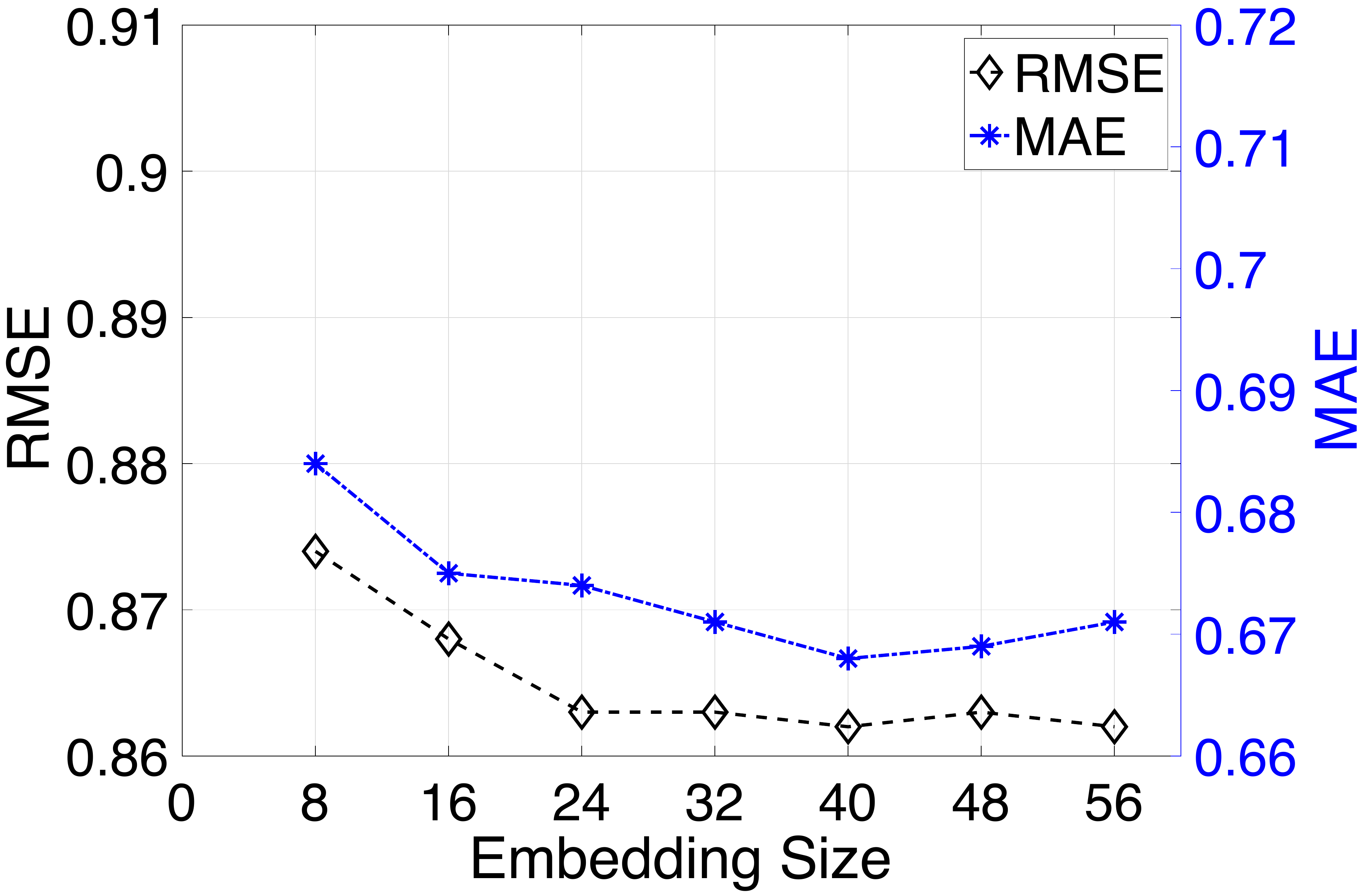}
        \label{fig:inf_para_embedding}
        }
    \subfigure[][\# of Hidden Layers]{
        \centering
        \includegraphics[width=0.22\textwidth]{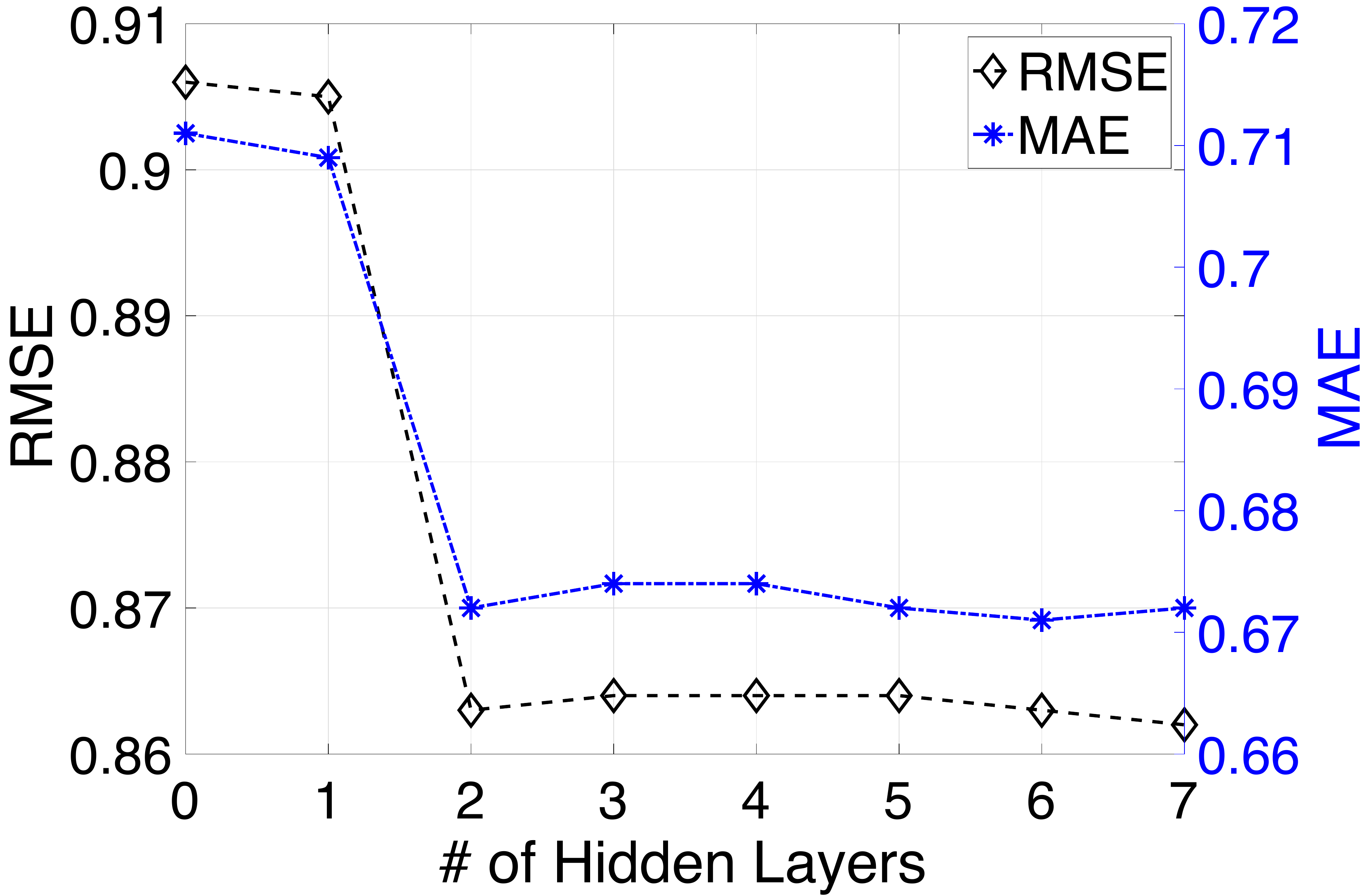}
        \label{fig:inf_para_layer}
        }
    \subfigure[][\# of Time Steps]{
        \centering
        \includegraphics[width=0.22\textwidth]{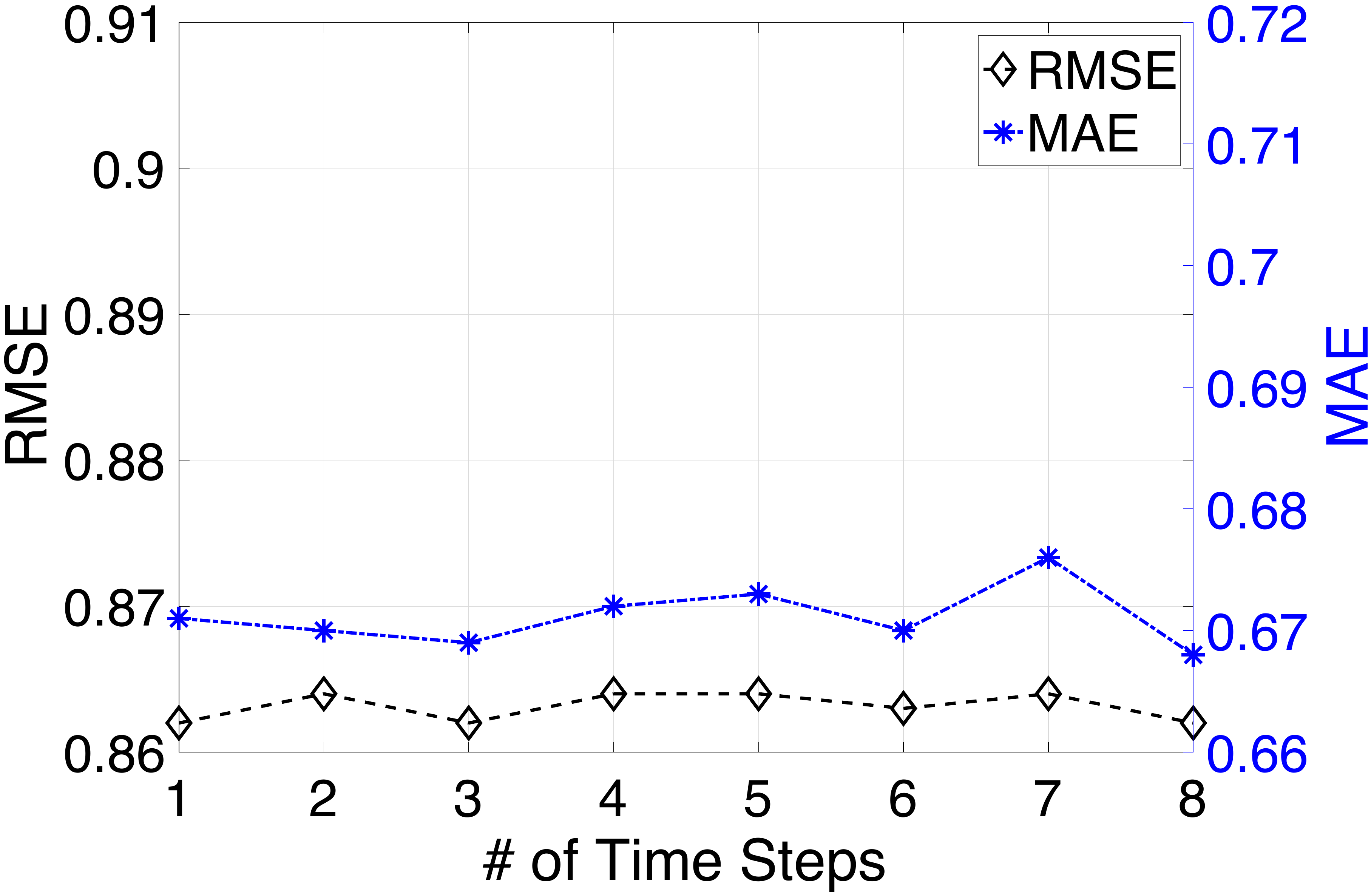}
        \label{fig:inf_para_sequence}
        }
    \subfigure[][Hidden State Dimension]{
        \centering
        \includegraphics[width=0.22\textwidth]{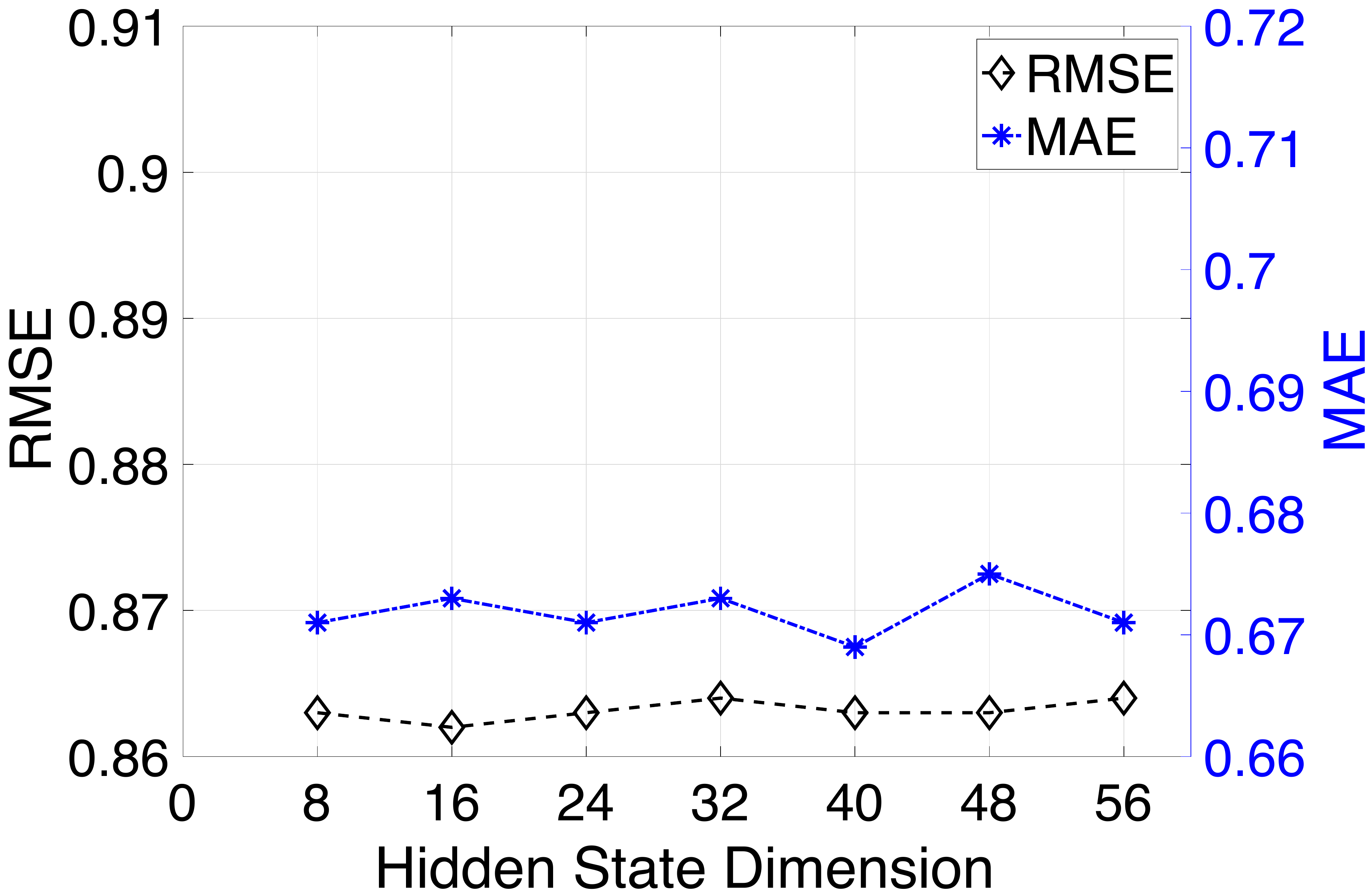}
        \label{fig:inf_para_state}
        }
    \vspace{-0.15in}
    \caption{Parameter sensitivity of \model($ReLU$) in rating prediction (50\% as training data, 10\% as validation data and the remaining as testing data).}
    \label{fig:para_sen_rating_infer}
    \vspace{-0.1in}
\end{figure*}


\begin{table}[t!]
\small
\centering
\renewcommand\arraystretch{1.2}
\vspace{-0.1in}
\caption{Performance of link prediction on Github archive.}
\vspace{-0.1in}
\begin{adjustbox}{max width=\linewidth}
\begin{tabular}{l| c c c c } 
\toprule
Metrics & AUC & F1-score  & Precision & Recall \\
\midrule
AA               & $0.5159$ & $0.1334$  & $0.4641$ & $0.0768$ \\
PA               & $0.6443$ & $0.3760$  & $0.5565$ & $0.2842$  \\
PMF              & $0.7278$ & $0.5233$  & $0.7507$ &$0.4013$ \\
BPMF             & $0.7165$ &$0.5291$  & $0.6978$ &$0.4250$  \\
NCF              & $0.7346$ & $0.5639$ & $0.6421$ & $0.5029$  \\
\midrule
\model $dot$   & $0.7133$ & $0.5328$ & $0.6681$ & $0.4437$  \\
\model($ReLU$)     &$0.7841$ &$0.6231$ &$0.6985$ &$0.5612$ \\
\model($sigmoid$)    &$0.7787$ &$0.6278$ &$0.6836$ &$0.5803$ \\
\model($tanh$)   &$0.7739$ &$0.6186$ &$0.6765$&$0.5689$ \\
\bottomrule
\end{tabular}
\end{adjustbox}
\label{tab:esti_link_pre}
\vspace{-0.15in}
\end{table}

\subsection{Rating Prediction and Inference (Q1, Q3 and Q4)}
We now compare \model\ with state-of-the-art techniques as we introduced above. To investigate the performance of all compared algorithms on different targeted time frames, we show the results from Jan, 2004 to Nov, 2004. The evaluation results are shown in Table~\ref{tab:esti_rating_time}. Furthermore, we provide analysis on the effects of training ratio (from 30\% to 80\%) on predictive performance, as shown in Table~\ref{tab:esti_rating_train}. Based on those evaluation results, we have the following four key observations. \\\vspace{-0.1 in}

\noindent (i) \textbf{Rating Prediction: Training/test time period}. We observe that \model\ consistently achieves the best performance over different time frames from Table~\ref{tab:esti_rating_time}. For example, \model\ achieves on average 0.075 and 0.076 (relatively 8.3\% and 10.7\%) improvements over TDSSM in terms of RMSE and MAE, and 0.075 and 0.075 (relatively 7.4\% and 7.8\%) improvements over RRN in terms of RMSE and MAE. The evaluation results across different time frames demonstrate the effectiveness of our \model\ framework in modeling time-evolving interactions between multiple dimensions in a dynamic scenario. Furthermore, the data becomes more dense as time window slides, \ie, density degree: 2004 Jan-3.67\%, 2004 Mar-3.86\%, 2004 May-4.11\%, 2004 Jul-4.34\%, 2004 Sep-4.45\% and 2004 Nov-4.57\%. We can observe that the performance gain between \model\ and other baselines become larger as data becomes sparser, suggesting our \model\ is capable of handing sparse relational data.\\\vspace{-0.1 in}

\noindent (ii) \textbf{Rating Inference: Training/test ratio}. Table~\ref{tab:esti_rating_train} shows the prediction results when varying the percentage of data in the training set. In this experiment, we fix the percentage of validation data as 10\% of the entire dataset. We can observe that obvious improvements can be obtained by our \model\ with different sizes of training data, demonstrating that \model\ is robust to the data sparsity issue. For example, the average relative improvement over on TDSSM and RRN algorithms are (RMSE: 5.5\%, MAE: 7.2\%) and (RMSE: 3.1\%, MAE: 3.3\%), respectively. In addition, we can observe rising trends as the training size increases, which indicate the positive effects of training data size on predicting ratings. Also, the performance gain between \model\ and other baselines becomes larger as the training size decreases (more sparse data), which validates the ability of \model\ model in predicting interactions with sparse tensor. We can notice that neural network based models (\ie, NCF and RRN) achieve better performance compared with conventional matrix factorization algorithms (\ie, PMF and BPMF) with less training data. This observation suggests that neural network based models are more suitable in sparse relational data. An interesting observation is that BPTF achieves better performance than TDSSM and RRN (neural methods), which indicates the effectiveness to consider the interactions between user, item and time embeddings. \\\vspace{-0.1 in}

\noindent (iii) \textbf{\model's variants}. We notice that the performance of our \model\ is not sensitive to different activation functions. In addition, our results also indicate that the necessity of our multi-layer perceptron component in \model\ for projecting inherent factors into a prediction output by capturing the non-linear relations between them.\\\vspace{-0.1 in}

\noindent (iv) \textbf{Performance gain analysis}. We can observe that \model\ shows improvement over both deep collaborative filtering based algorithms (\ie, NCF, PMF, BPMF and BPTF) and recurrent neural networks based schemes (\ie, RRN and TDSSM). In particular, \emph{Firstly}, this sheds light on the limitations of collaborative filtering based algorithms which ignore the temporal dynamics among the multi-dimensional interactions in the rating data. \emph{Secondly}, the large performance gap between \model\ and recurrent neural network based schemes indicates the limitation of those approaches which only model the sequential pattern of the tensor's temporal dimension and fail to consider the dependencies between the implicit interactions across dimensions.

\begin{figure*}[thb!]
    \centering
    \vspace{-0.1 in}
    \subfigure[][Embedding Size]{
        \centering
        \includegraphics[width=0.22\textwidth]{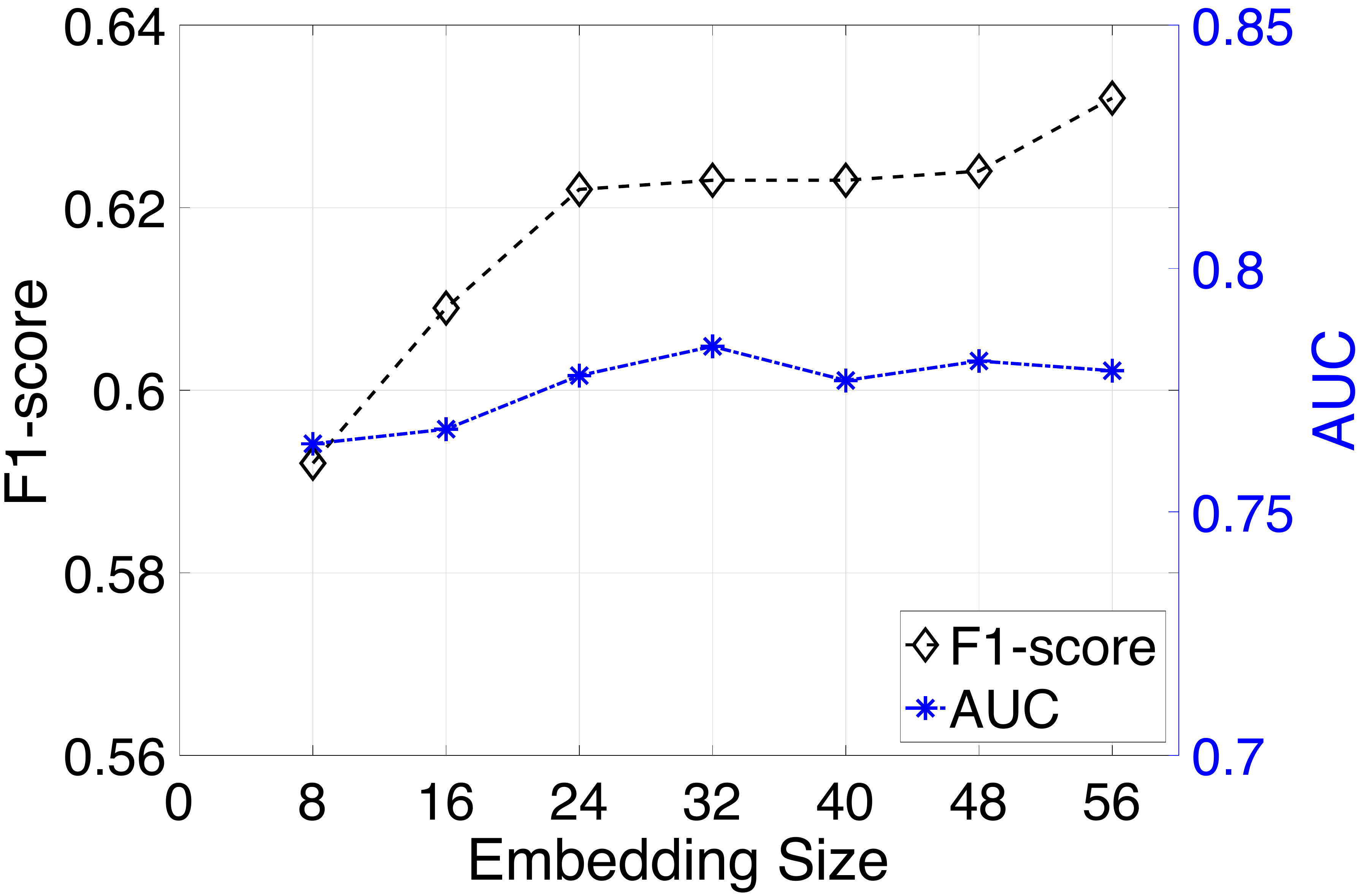}
        \label{fig:link_para_embedding}
        }
    \subfigure[][\# of Hidden Layers]{
        \centering
        \includegraphics[width=0.22\textwidth]{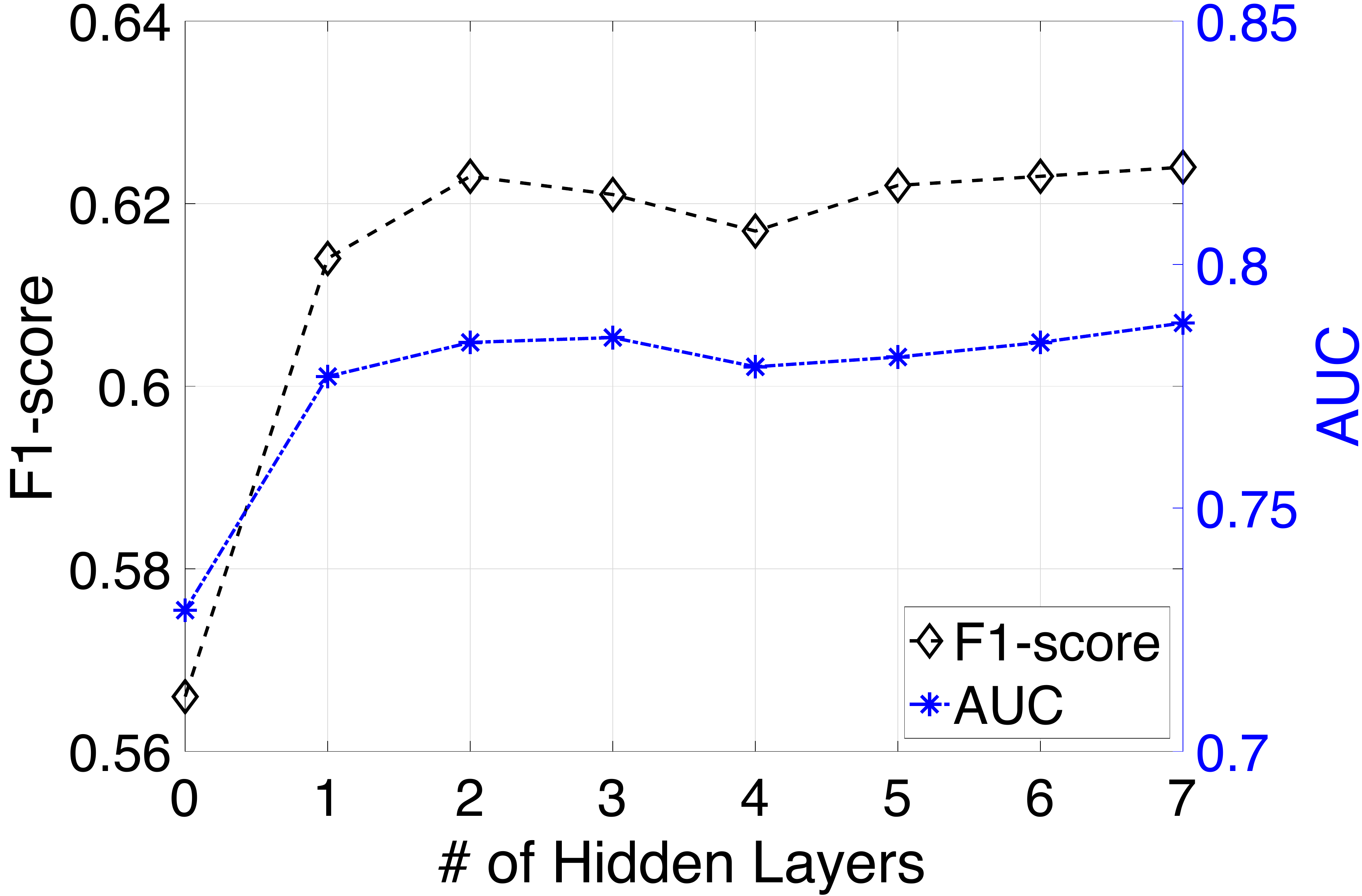}
        \label{fig:link_para_layer}
        }
    \subfigure[][\# of Time Steps]{
        \centering
        \includegraphics[width=0.22\textwidth]{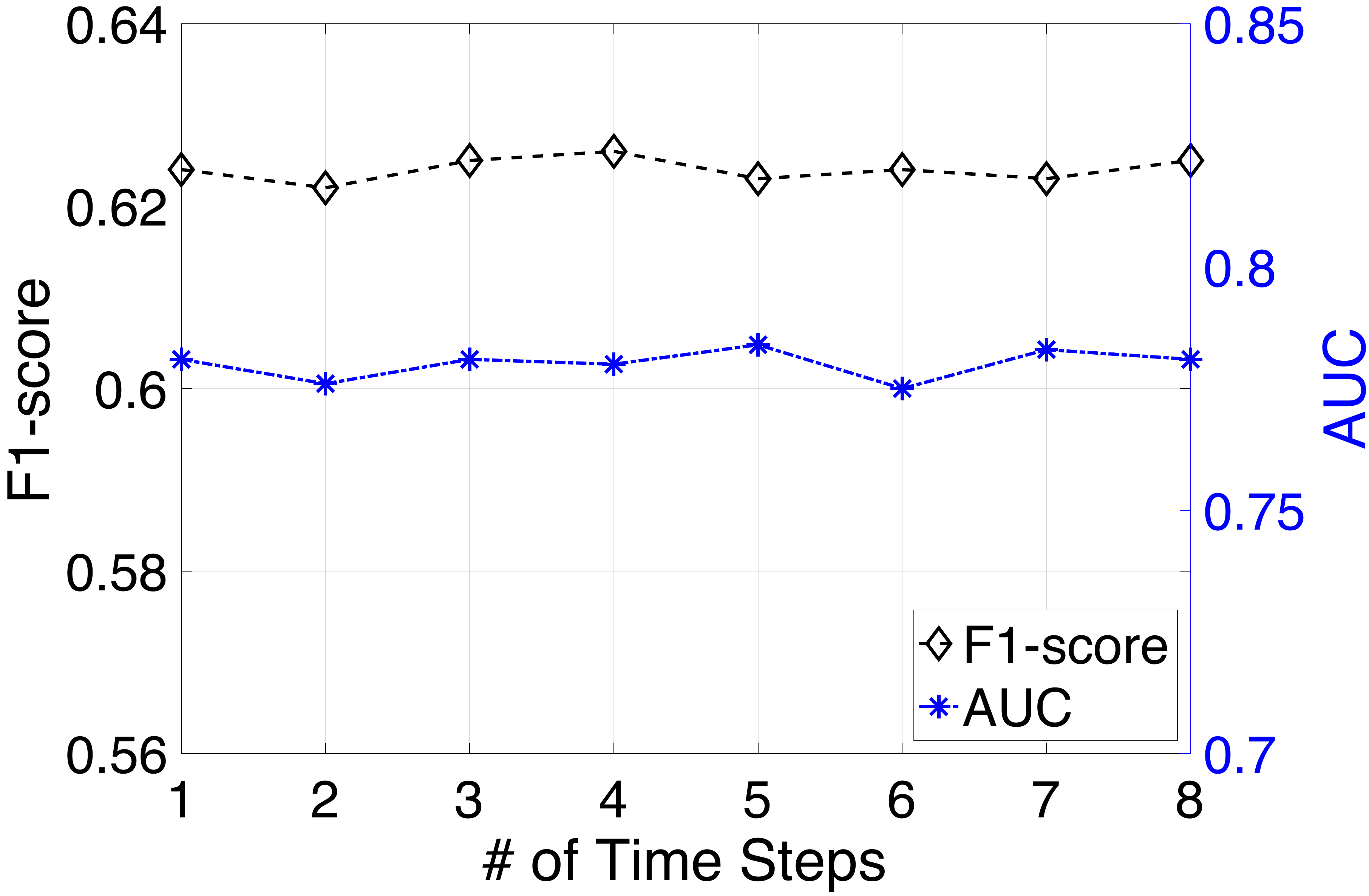}
        \label{fig:link_para_sequence}
        }
    \subfigure[][Hidden State Dimension]{
        \centering
        \includegraphics[width=0.22\textwidth]{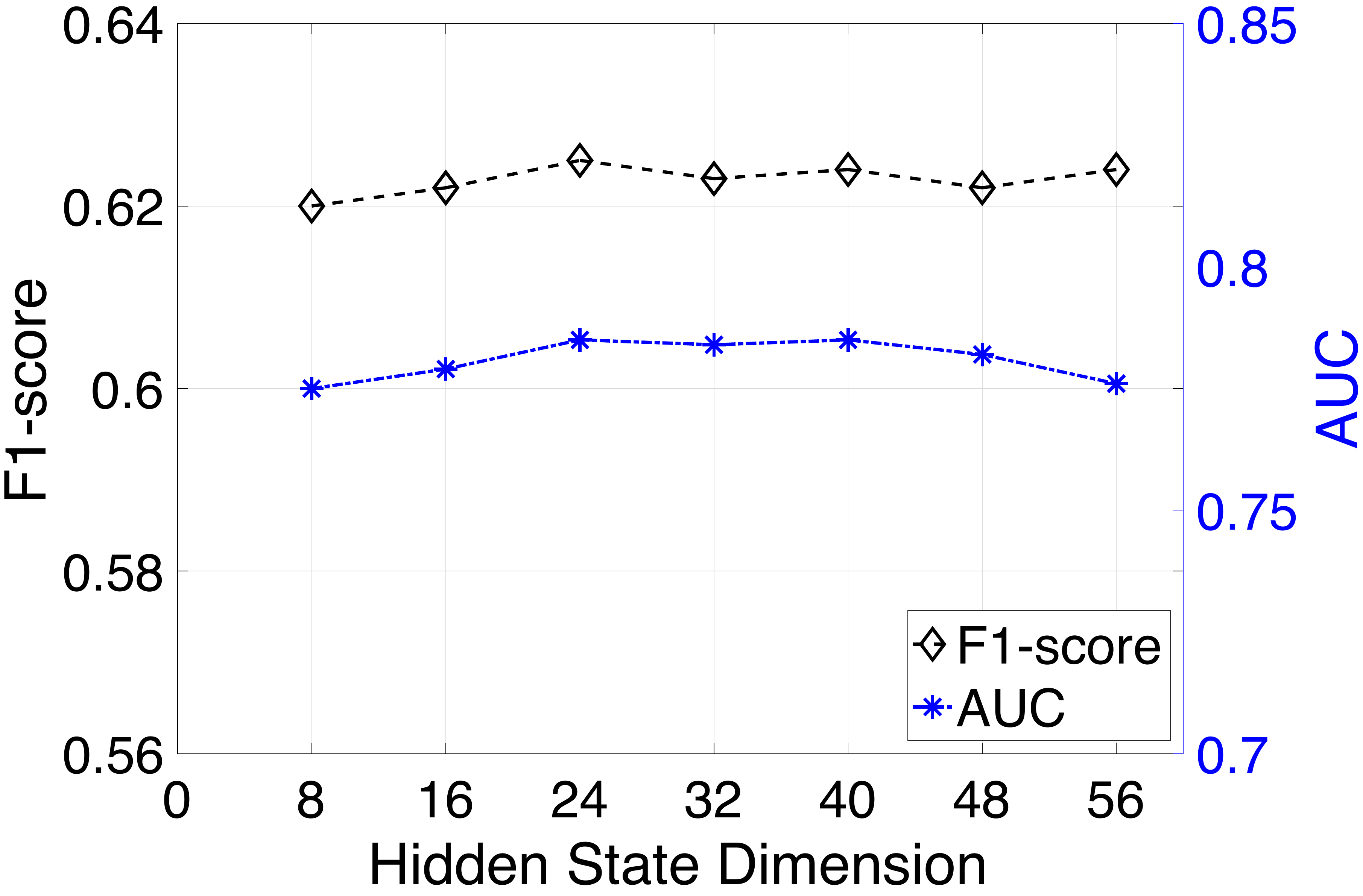}
        \label{fig:link_para_state}
        }
    \vspace{-0.15in}
    \caption{Parameter sensitivity of \model($ReLU$) in link prediction on Github dataset.}
    \label{fig:para_sen_link_pre}
    \vspace{-0.1in}
\end{figure*}

\subsection{Link Prediction (Q2, Q3 and Q4)}
Table~\ref{tab:esti_link_pre} lists the evaluation results of the link prediction task. In this evaluation, we evaluate all compared models using the the 90\% and 10\% of Github archive data from first twenty weeks as training and validation data to predict links in the twenty-first week. To construct the testing set, we use the observed links in the twenty-first week as the positive cases and randomly enumerate node pairs and choose unobserved edges as negative cases. We use the same parameter settings as rating prediction task (listed in Table~\ref{tab:para_setting}). Furthermore, because link prediction problem often suffers from highly unbalanced data (\ie, only 0.02\% instances are observed in the GitHub dataset), we sample the negative cases twice as many as positives cases to address this challenge following~\cite{chawla2004special}. 



Overall, the proposed \model\ significantly outperforms other baselines in F1 and AUC. Specifically, the relative improvement of our \model\ over NCF is 10.2\% and 6.8\% in terms of F1-score and AUC respectively. This link prediction task further demonstrate that the \model\ framework works well by capturing time-evolving interactions between different dimensions in a non-linear manner.


\subsection{Parameter Sensitivity (Q5)}
The \model\ model involves several parameters (\ie, \emph{embedding size} in embedding layer, \emph{\# of hidden layers} in MLP, \emph{\# of time steps} and \emph{hidden state dimension} in LSTM). To investigate the robustness of \model\ framework, we examine how the different choices of parameters affect the performance of \model\ in both rating and link prediction tasks. Except for the parameter being tested, we set other parameters at the default values (see Table~\ref{tab:para_setting}).

\noindent \textbf{Rating Prediction}.
Figure~\ref{fig:para_sen_rating_infer} and Figure~\ref{fig:para_sen_rating_pre} list the prediction results (measured by RMSE and MAE) as a function of one selected parameter when fixing others. Note that we have two $y$-axes corresponding to RMSE (left-black) and MAE (right-blue) respectively due to their different value ranges.
From Figure~\ref{fig:para_sen_rating_infer}, overall, we can observe that \model\ is not strictly sensitive to these parameters, except for \# of hidden layers, and can achieve high performance with cost-effective parameters, \ie, the smaller the parameters are, the more efficient the training process will be. Figure~\ref{fig:inf_para_layer} and Figure~\ref{fig:pre_para_layer} indicate that model performance becomes stable as long as the number of hidden layers is above 2. From Figure~\ref{fig:inf_para_embedding}, we can observe that embedding size is positively correlated with the prediction accuracy and we set it to 32 in our experiment due to the balance between efficacy and computational cost. Additionally, we can observe the low impact of other two parameters (\ie, sequence length and state size in LSTM) on model performance, which suggests the robustness of our \model in modeling the temporal dynamics of multi-dimensional interactions.

\noindent \textbf{Link Prediction}.
We also study the parameter sensitivity of \model\ as measured by link prediction performance. Figure~\ref{fig:para_sen_link_pre} shows the prediction accuracy (measured by F1-score and AUC) as a function of each of the four parameters when fixing the other three. Figure~\ref{fig:link_para_sequence} and Figure~\ref{fig:link_para_state} suggest that the sequence length and state size have little impact on prediction accuracy. The increase of link prediction performance converges as the number of hidden layers reaches around 4. Additionally, we can observe that our model shows an increasing trend with an increasing embedding size from Figure~\ref{fig:link_para_embedding}, which is consistent with the observation from the parameter sensitivity evaluation in rating prediction. For \model\ without hidden layer (final predictions are directly derived from embedding layer), the performance is suboptimal. This observation verifies our argument that the dot product operation cannot handle the non-linear interactions in tensor factorization and demonstrates the necessity to model the complex interaction dependencies with hidden layers.

\section{Related Work}
\label{sec:relate}

\noindent \textbf{Deep Collaborative Filtering Models.} 
Collaborative Filtering (CF) has been widely applied to various recommendation systems~\cite{he2017neural,wu2016collaborative,li2017collaborative,hsieh2017collaborative,chen2017attentive}. In particular, He \etal\ aimed to develop a neural network collaborative filtering framework by modeling latent features of users and items~\cite{he2017neural}. Wu \etal\ studied the top-N recommendation problem and proposed a autoencoder based CF method~\cite{wu2016collaborative}. Furthermore, a collaborative variational autoencoder has been developed in recommendation systems to consider implicit relationships between items and users~\cite{li2017collaborative}. Hsieh \etal\ studied the connection between metric learning and collaborative filtering. However, these approaches are static models and are lacking when they comes to dynamic scenarios. The proposed \model\ addresses this problem by modeling temporal evolution of latent factors in collaborative filtering framework.

\noindent \textbf{Deep Matrix Factorization.} With the advent of deep learning techniques, significant effort has been made to develop neural network-based matrix factorization models~\cite{sedhain2015autorec,sainath2013low,kim2016convolutional,dziugaite2015neural}. Sedhain \etal\ proposed an autoencoder framework for collaborative filtering~\cite{sedhain2015autorec}. Sainath \etal\ proposed to apply low-rank factorization to deep neural network models to address the language modeling problem~\cite{sainath2013low}. More recently, to address the sparsity problem in recommendation techniques, Kim \etal\ designed a model which integrates convolutional neural network (CNN) into probabilistic matrix factorization (PMF)~\cite{kim2016convolutional}. Dziugaite \etal\ suggested to replace the the inner product in matrix factorization with the function which is learned from the data together with latent feature vectors. However, the limitation of the above approaches is that they only consider static data instead of dynamic data in which temporal dimension need to be explored. Our work furthers the investigation on this direction by developing the \model\ framework to capture the time-evolving temporal dynamics exhibited from relational data with multiple types of entity dependencies, which cannot be handled by previous models.

\noindent \textbf{Applications of Matrix Factorization}. There is a good amount of work on the applications of Matrix Factorization. Existing recommendation techniques can be grouped into three categories: content-based algorithms~\cite{cantador2010content,van2013deep}, collaborative filtering based algorithms~\cite{mnih2008probabilistic,bu2016improving} and hybrid algorithms~\cite{he2016ups,wang2015collaborative}. For example, several content-based recommendation models have been evaluated based on the profiles of users and items~\cite{cantador2010content}. Salakhutdinov \etal\ presented a Probabilistic Matrix Factorization (PMF) model and demonstrated its effectiveness on the movie rating data~\cite{mnih2008probabilistic}. Additionally, Wang \etal\ proposed a hierarchical Bayesian model which integrated content information and collaborative filtering scheme by performing deep representation learning~\cite{wang2015collaborative}. Li \etal\ generalized latent factor framework for social network analysis by modeling homophily. This work can be complementary to the above works in the sense that explicitly exploring temporal dynamics in relational data normally lead to better recommendation results.
\section{Conclusion}
\label{sec:conclusion}

We developed a novel and general \full\ (\model) for modeling dynamic relational data that addresses the critical challenge of evolving user-item relational data. By modeling the time-evolving inherent factors and incorporating temporal smoothness constraints on those factors, \model\ is capable of capturing both the time-varying interactions across dimensions and the non-linear relations between them. Extensive experiments on two real-world datasets in rating prediction and link prediction tasks show that \model significantly outperforms baseline methods.

Notwithstanding the interesting problem and promising results, some directions exist for future work. We will next incorporate rich heterogeneous auxiliary data to further improve the model. Another possible direction is adapting~\model~to a time-sensitive model by analyzing the trade-off between accuracy and complexity.



\bibliographystyle{ACM-Reference-Format}
\bibliography{sigproc} 

\end{document}